\documentclass[journal]{IEEEtran}

\usepackage[utf8]{inputenc} % allow utf-8 input
\usepackage[T1]{fontenc}    % use 8-bit T1 fonts
\usepackage{url}            % simple URL typesetting
\usepackage{booktabs}       % professional-quality tables
\usepackage{amsfonts}       % blackboard math symbols
\usepackage{nicefrac}       % compact symbols for 1/2, etc.
\usepackage{microtype}      % 
\usepackage{enumitem}
\usepackage{algorithm,algpseudocode}
\usepackage{amsmath,amsthm,amssymb,mathtools,enumerate,array}
\usepackage{graphicx,tabularx,epstopdf,url}
\usepackage{cite}

\makeatletter
\g@addto@macro\normalsize{%
  \setlength\abovedisplayskip{2pt}
  \setlength\belowdisplayskip{2pt}
  \setlength\abovedisplayshortskip{1pt}
  \setlength\belowdisplayshortskip{1pt}
}
\makeatother

\usepackage[skip=2mm]{caption}
\usepackage{xcolor}
\usepackage{hyperref}       % hyperlinks
\hypersetup{
    colorlinks,
    linkcolor={red!80!black},
    citecolor={blue!50!black},
    urlcolor={blue!90!black}
}

\def\ba{\mathbf{a}}

\def\bb{\mathbf{b}}

\def\bp{\mathbf{p}}
\def\bs{\mathbf{s}}
\def\bx{\mathbf{x}}
\def\by{\mathbf{y}}
\def\bz{\mathbf{z}}

\def\btheta{\boldsymbol{\theta}}
\def\bphi{\boldsymbol{\phi}}
\def\bPhi{\boldsymbol{\Phi}}
\def\Phibar{\overline{\boldsymbol{\Phi}}}

\def\bmu{\boldsymbol{\mu}}

\def\rone{\overrightarrow{\mathbf{1}}}

\def\bbR{\mathbb{R}}
\def\bbE{\mathbb{E}}

\def\bI{\mathbf{I}}
\def\bA{\mathbf{A}}
\def\bB{\mathbf{B}}
\def\bC{\mathbf{C}}

\def\bM{\mathbf{M}}

\def\bP{\mathbf{P}}
\def\bS{\mathbf{S}}
\def\bU{\mathbf{U}}
\def\bV{\mathbf{V}}
\def\bW{\mathbf{W}}
\def\bX{\mathbf{X}}
\def\bY{\mathbf{Y}}
\def\bZ{\mathbf{Z}}

\def\bone{\mathbf{1}}

\def\Sb{{\mathbf{S}_B}}
\def\bK{\mathbf{K}}
\def\Kbar{\bar{\mathbf{K}}}
\def\Kb{{\mathbf{K}_B}}

\def\Sbar{\bar{\mathbf{S}}}
\def\Xbar{\bar{\mathbf{X}}}
\def\bSigma{\boldsymbol{\Sigma}}

\def\lnorm{\left\|}
\def\rnorm{\right\|}
\def\lp{\left(}
\def\rp{\right)}

\DeclareMathOperator{\trace}{tr}

\theoremstyle{plain}

\newtheorem{definition}{Definition}

\usepackage{color}

\newcolumntype{Y}{>{\centering\arraybackslash}X}

\newcommand{\PreserveBackslash}[1]{\let\temp=\\#1\let\\=\temp}
\newcolumntype{C}[1]{>{\PreserveBackslash\centering}p{#1}}

\title{Privacy Enhancing Machine Learning via Removal of Unwanted Dependencies}
\author{
  Mert Al\textsuperscript{1},\quad Semih Yagli\textsuperscript{2},\quad Sun-Yuan Kung\textsuperscript{2}%\thanks{Use footnote for providing further information about author (webpage, alternative address)---\emph{not} for acknowledging funding agencies.} 
  \\
  Princeton University \\
  \texttt{\textsuperscript{1}merta@alumni.princeton.edu, \textsuperscript{2}syagli@princeton.edu, \textsuperscript{3}kung@princeton.edu} \\
}

\allowdisplaybreaks
\begin{document}
% use for special paper notices
%\IEEEspecialpapernotice{(Invited Paper)}

% make the title area
\maketitle

% As a general rule, do not put math, special symbols or citations
% in the abstract or keywords.
\begin{abstract}
The rapid rise of IoT and Big Data has facilitated copious data driven applications to enhance our quality of life. However, the omnipresent and all-encompassing nature of the data collection can generate privacy concerns. Hence, there is a strong need to develop techniques that ensure the data serve only the intended purposes, giving users control over the information they share. To this end, this paper studies new variants of supervised and adversarial learning methods, which remove the sensitive information in the data before they are sent out for a particular application. The explored methods optimize privacy preserving feature mappings and predictive models simultaneously in an end-to-end fashion. Additionally, the models are built with an emphasis on placing little computational burden on the user side so that the data can be desensitized on device in a cheap manner. Experimental results on mobile sensing and face datasets demonstrate that our models can successfully maintain the utility performances of predictive models while causing sensitive predictions to perform poorly.
\end{abstract}

% Note that keywords are not normally used for peerreview papers.
\begin{IEEEkeywords}
Data privacy, Adversarial learning, Representation learning, Kernel methods, Dimension reduction.
\end{IEEEkeywords}

% For peer review papers, you can put extra information on the cover
% page as needed:
% \ifCLASSOPTIONpeerreview
% \begin{center} \bfseries EDICS Category: 3-BBND \end{center}
% \fi
%
% For peerreview papers, this IEEEtran command inserts a page break and
% creates the second title. It will be ignored for other modes.
\IEEEpeerreviewmaketitle

\section{Introduction}

With more of our daily activities moving online, a vast amount of personal information is being collected, stored and shared across the internet. Although this information can be used for the benefit of the data owners, it can also leak sensitive information about individuals. Mobile-sensing readings, for instance, can be beneficially used for activity recognition \cite{kwapisz2011activity}, medical diagnosis \cite{agu2013smartphone}, or authentication \cite{kwapisz2010cell}; nevertheless, they can also be used to infer sensitive information about individuals such as location, context and identity \cite{narain2016inferring,christin2011survey}. 

The possibility of applying machine learning (ML) for adversarial purposes motivates the application of the principle of least privilege to big data \cite{saltzer1975protection}, i.e., to give service providers access to only the information necessary for the intended utility, but nothing else. Our methods, hence, follow this principle by seeking the feature representation of the data such that it maximizes the information on the utility task, but removes unwanted correlations.

\begin{figure}[t]
  \begin{minipage}[b]{\linewidth}
   \centering
   \centerline{\includegraphics[trim = 30 70 30 70,clip,width=\linewidth]{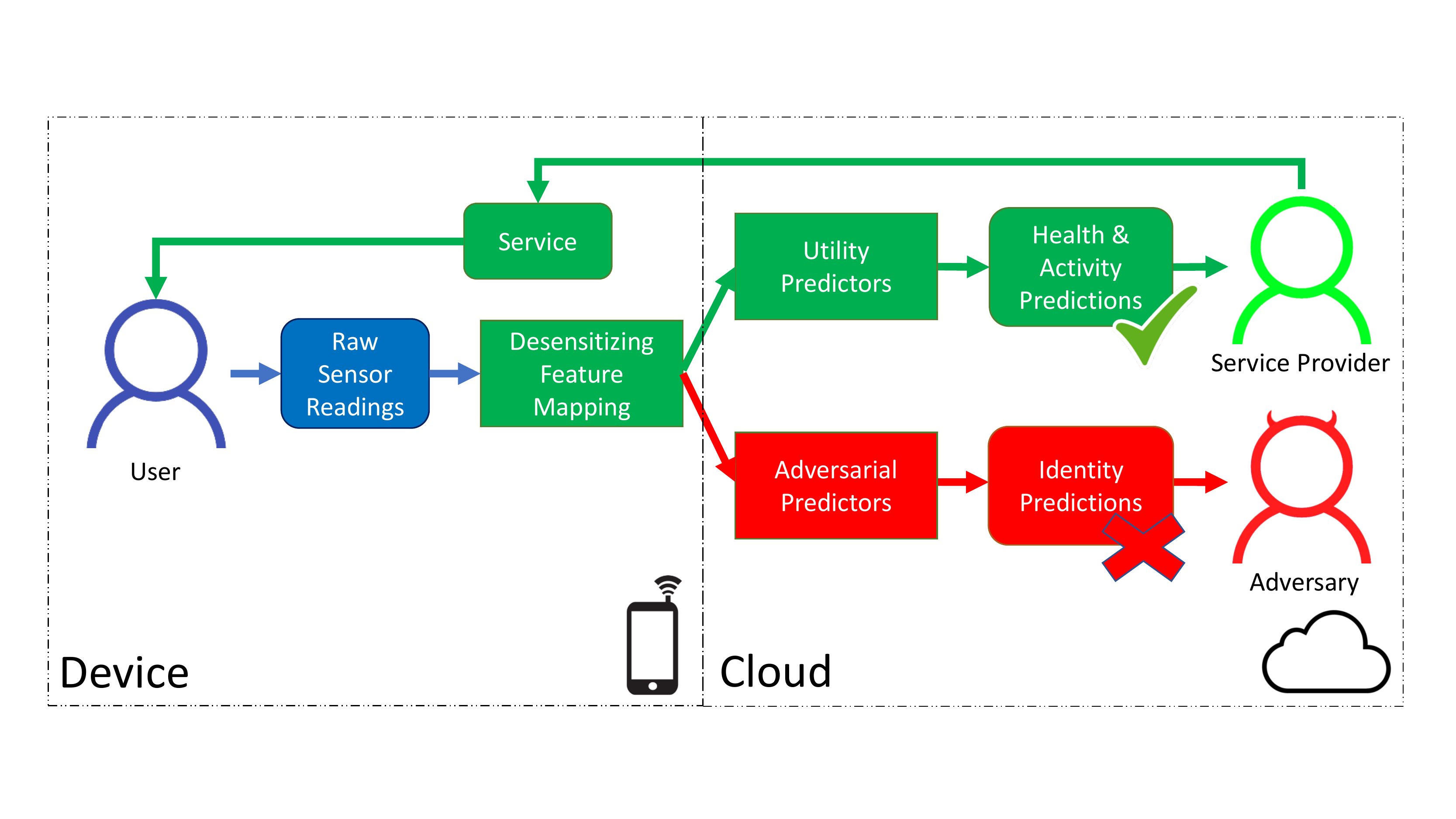}}
  \end{minipage}
  \caption[Illustration of our learning setting]{A privacy sensitive learning setting separated between the private sphere (device) and public sphere (cloud). The desensitizing feature mapping removes identity related correlations from the raw sensor readings while maintaining the activity related correlations, resulting in high predictive performance for the service provider and low performance for the adversary.}
  \label{fig:BigPicture}
\end{figure}

Our work is intended to allow privacy preservation to be performed by the data owner even before any information can be extracted. Therefore, we consider two spheres in our design, the \emph{private sphere} and the \emph{public sphere} as illustrated by Figure \ref{fig:BigPicture}. Based on this separation, \emph{lossy compression needs to occur in the private sphere, such that any data released to the public sphere should be viable only to the intended purpose}. To achieve such design, we employ sequential models, whose computations can be seamlessly divided between private and public spheres. From the system's perspective, the private sphere is thus concerned with employing data compression to maximize the utility information while removing redundant or sensitive information. The public sphere, on the other hand, is concerned with making utility predictions on the compressed data. Since both spheres serve the same utility goal, we optimize them jointly. Also, in an effort to reduce the burden on the users, we place the majority of the model computations in the public sphere, i.e., after the data is desensitized by the owner.

To identify and mitigate the unwanted correlations in the data, our methods need to have access to samples that may contain sensitive information. For this reason, we initially consider ourselves in a setting, where either public data are available for the training of the privacy enhancing models, or there is a trusted third party with access to private data, who can train privacy enhancing models before they are deployed. Restriction to these scenarios can be alleviated by privacy preserving collaborative learning methods such as homomorphic encryption and Differential Privacy. Moreover, it is possible for users to avoid sending their data outside by training and combining local models via federated learning. However, these extensions are beyond the scope of this paper.

The main contributions in this paper are summarized below. We hope that these advances will motivate future research into this under-explored learning setting with conflicting utility and privacy goals. 
\begin{itemize}
    \item We build and explore a variety of optimization objectives tailored towards removing dependencies between data representations and sensitive attributes. Among these, the Maximum Mean Discrepancy (MMD) \cite{li2015generative}, the Wasserstein Discriminator Network (WDN) \cite{arjovsky2017wasserstein} and the Least Squares Discriminator Network (LSDN) \cite{mao2017least} objectives have their roots in the generative adversarial learning framework, while the Kernel Discriminant Information (KDI) \cite{al2020scalable} objective has its roots in Kernel Discriminant Analysis. Because these objectives have not been used in scenarios similar to ours, we make substantial efforts to analyze their merits relative to each other.
    \item We present novel techniques for optimizing privacy preserving feature mappings as well as the predictive models that use them as inputs. Our methods are lightweight in the sense that they require minimal modifications to existing learning algorithms and model structures.
    \item We demonstrate the viability of our privacy enhancing learning techniques on mobile sensing and face image datasets, where we hide the identities of the users. Our experiments showcase that, under the right conditions, we can remove almost all the sensitive information within the data with minimal loss in utility performance. Furthermore, our methods are able to limit the utility performance losses in high privacy settings even when we restrict ourselves to linear feature mappings on the user side.
\end{itemize}

\subsection{Network Architectures}

\subsection{Related Works}

A large body of work onto privacy enhancing machine learning focuses on making the model parameters or predictions differentially private with respect to their training examples \cite{xu2015privacy,mohassel2017secureml,chaudhuri2011differentially,liu2015fast,yu2019differentially,chaudhuri2013stability,kusner2015differentially}. That is to say, these works make the models statistically indistinguishable when conditioned on the presence or absence of a training sample, effectively hiding individual samples among a crowd of training data. Our work is orthogonal to this approach, because we aim to remove unwanted information directly from individual samples. Nevertheless, our approach could be combined with differentially private learning mechanisms to hide the participation of users, when their data serve to train the privacy preserving models that we develop.

Some of the early works concerned with desensitizing data rely on random projections to preserve pairwise distances between data points, while making the original data entries difficult to reconstruct \cite{liu2006random,liu2012cloud}. While such projections may serve some privacy benefits and maintain the performance of certain ML models, they make no distinctions between desirable and undesirable correlations in the data, hence, they are insufficient to serve our goals. More recent works in \cite{kung2017compressive,kung2017compressive2}, on the other hand, propose suitable projection techniques, which maximize correlations with a utility variable and minimize correlations with a privacy variable. These works utilize the linear variant of the Discriminant Information (DI) criterion (a more general criterion is covered in Section \ref{sec:kdi}) for both the utility and the privacy targets, though, they do not go beyond optimizing shallow linear or kernel based projections of the data. 

The work in \cite{al2018multi} separates learning models into public and private spheres much like our design, but, the compression methods in the private sphere only focus on the utility goals and do not take explicit privacy targets into account. In \cite{louizos2015variational}, the authors attempt to remove dependencies between binary sensitive variables and dense neural network representations by utilizing an approximation of the Maximum Mean Discrepancy (MMD) objective (covered in Section \ref{sec:mmd}). In \cite{feng2019learning} an alternative objective is proposed via linear mappings approximating the Wasserstein Distance (a more general methodology is covered in Section \ref{sec:wdn}) and a multi-class extension is included similar to the one we present in Section \ref{sec:multi-class}. Adversarial networks are also employed to remove unwanted sources of variation in \cite{ganin2016domain,xie2017controllable,jaiswal2020invariant} for tasks such as domain adaptation, lighting independent image classification and achieving fair model predictions.

Our methods are suitable for optimizing more general classes of data representations than the previous works, such as convolutional neural network (CNN) mappings, which, as we showcase, are not appropriate for our purposes without some modifications. Additionally, we present and analyze a comprehensive set of learning objectives suitable for removing unwanted dependencies in the data. The proposed objectives are capable of handling multi-class private variables, with some being capable of handling continuous-valued private variables as well.

\section{Privacy Enhancing Learning Objectives} \label{sec:privacy_objectives}

\subsection{Notation}

Throughout this paper, we refer to (deterministic) matrices with bold, capital letters, vectors with bold, small letters and scalars with regular letters. We reserve the letter $X$ to refer to data, $Z \coloneqq \phi(X)$ to refer to the \emph{processed data} and $S$ to refer to sensitive information. When we talk about empirical data, $\bX=[\bx_1\ldots\bx_N]$ refers to the data matrix, $\bZ=[\bz_1\ldots\bz_N] \coloneqq [\phi(\bx_1)\ldots\phi(\bx_N)]$ refers to the \emph{processed data matrix}, and $\bs = [s_1 \ldots s_N]^\top$ refers to the privacy label vector containing the sensitive attributes. If we talk about privacy labels in matrix form, we refer to $\bP=[\bs_1 \ldots \bs_N]^\top$. Because the letter $\bS$ is commonly used to denote the scatter matrix, we reserve it for that purpose. We commonly use $\bC \coloneqq \bI-\frac{1}{N}\rone\rone^\top$ to refer to the centering matrix. Also, for a feature mapping $\phi(\cdot)$ that takes as input a data sample $\bx$, we commonly use $\phi(\bX)$ as a shorthand to refer to $[\phi(\bx_1)\ldots \phi(\bx_N)]$.

We generally refer to kernel functions as $k(\cdot,\cdot)$, and reserve the letter $\bK$ for the $N \times N$ kernel matrix obtained from the processed data matrix $\bZ$, where $\bK_{ij}=k(\bz_i,\bz_j)$. Additionally, we use $k(\bA,\bB)$ as a shorthand to refer to a kernel matrix with $k_{ij}(\bA,\bB)=k(\ba_i,\bb_j)$.

$\|\cdot\|_2$ refers to the $l_2$ norm for vectors and the spectral norm for matrices, while $\|\cdot\|_F$ refers to the Frobenius norm. For a matrix $\bM$, we use $\trace(\bM)$ to denote its trace, $\bM^{-1}$ to denote its inverse, $\bM^{+}$ to denote its pseudo-inverse. Finally, when we talk about a loss function $L(\cdot;\cdot)$, the parameters used to minimize the loss come after the semi-colon, and anything before is treated as a constant.

\subsection{Background}\label{subsec:background}

In this section, we discuss learning objectives that can be useful for removing sensitive information from data. Before we delve into details, we first ask the fundamental question: Given only empirical data and no prior information about the underlying structure, how can we characterize the level of privacy achieved by a system? 

Obviously, perfect privacy is achieved in the case when the processed data $Z=\phi(X)$ is independent from the sensitive attribute $S$. We can characterize this system with  $P_{ZS}=P_Z P_S$, or $P_{Z|S}=P_Z$, that is, the distribution of the processed data is not affected by any realization of the sensitive attribute $S$. Since we do not know the underlying distribution $P_{XS}$, we need to measure the significance of the difference between $P_Z$ and $P_{Z|S}$ purely based on an i.i.d. sample $\{(\bx_i,\bs_i)\}_{i=1}^{N}$ with $(\bx_i,\bs_i) \sim P_{XS}$.

For simplicity, let us consider the binary case $S \in \{0,1\}$. Then, perfect privacy corresponds to the realization of $P_{Z|S=0}=P_{Z|S=1}$, where $Z=\phi(X)$. Given only a finite sample generated by $P_{XS}$, an arbitrary mapping $\phi$ and no knowledge of the data generating process, however, we cannot determine whether this system achieves perfect privacy without any doubt. What we can do is to test our confidence in the hypothesis $P_{Z|S=0}=P_{Z|S=1}$ based on our observations. Thus, a measure of privacy based solely upon empirical data can be defined as follows.

\begin{definition}[Model Free Binary Privacy Measure]
Let $D(\mathcal{S}_0,\mathcal{S}_1)$ be a measure of distance between the empirical distributions of two sets of samples $\mathcal{S}_0$, $\mathcal{S}_1$. Assume we have $N$ i.i.d. samples $\{(\bz_i,s_i)\}_{i=1}^{N}$, each obtained from $P_{ZS}$, with $s_i\in \{0, 1\}$. Let $D({\mathcal{Z}_0},{\mathcal{Z}_1})=\gamma$ be the observed distance between the empirical distributions of two disjoint subsets of $\{\bz_i\}_{i=1}^{N}$: $\mathcal{Z}_0=\{\bz_i\colon s_i=0\}$, $\mathcal{Z}_1=\{\bz_i\colon s_i=1\}$, with $|\mathcal{Z}_j|=N_j$. Then, our privacy confidence is given by $\mathbb{P}[D({Z_0},{Z_1}) \le \gamma]$, where $Z_0$, $Z_1$ are independent random sets of samples with sizes $N_0$ and $N_1$, which are obtained i.i.d. from the marginal distribution $P_Z$.
\label{def:model_free_priv}
\end{definition}

We basically defined our privacy measure as our confidence in the null hypothesis: \textit{The processed data $Z=\phi(X)$ is independent from the binary sensitive attribute $S$}. Accordingly, we assume that $P_{Z|S=0}=P_{Z|S=1}=P_Z$, then obtain the likelihood of the observed difference between empirical distributions. 

While the actual methods to obtain $p$-values for such hypotheses can be quite involved (often incorporating density estimation methods and bootstrapping), all we need to observe from Definition \ref{def:model_free_priv} is that \emph{the smaller $D({\mathcal{Z}_0},{\mathcal{Z}_1})$ is, the more confident we are in the assumption that $Z$ does not leak information about $S$}. Hence, for a parametric feature mapping $\phi(\bx;\btheta)$, an appropriate privacy preservation objective would be $D(\{\phi(\bx_i;\btheta)\colon s_i=0\},\{\phi(\bx_i;\btheta)\colon s_i=1\})$. 

To summarize, \emph{an appropriate empirical privacy objective tries to make the distributions of processed data look identical when conditioned on different values of the private variable}. While the example we gave works for binary private variables only, we can generalize it to multi-class private variables in a straightforward \textit{One-vs-Rest} fashion. Some of the objectives we shall present are also straightforward to generalize to continuous private variables, as they are related to the minimum least-squares error.

\subsection{Integral Probability Metrics} \label{sec:ipms}

We covered how privacy objectives can be defined from distances between empirical distributions in Section \ref{subsec:background}, now, we shall introduce a class of distance measures that are suitable for empirical data. Integral Probability Metrics (IPMs) are among the most commonly used distance measures in the literature to ensure closeness of empirical distributions \cite{gretton2012kernel,li2015generative,li2017mmd,arjovsky2017wasserstein}. These can be generally defined as follows.

\begin{definition}[Integral Probability Metric]
Let $P_0$, $P_1$ be two probability measures defined on $\mathcal{X}$, with $Z_0 \sim P_0$ and $Z_1 \sim P_1$. Let $\mathcal{F}$ be a class of bounded functions $f\colon \mathcal{Z} \rightarrow \bbR$, an Integral Probability Metric $D(\cdot,\cdot)$ is defined as
\begin{equation}
D_{IPM}(P_0,P_1)=\underset{f \in \mathcal{F}}{\sup} \left\{ \bbE\left[f(Z_0)\right]-\bbE\left[f(Z_1)\right] \right\} \text{.}
\label{eq:IPM}
\end{equation}
\label{def:IPM}
\end{definition}

To define a proper metric, we require the function class $\mathcal{F}$ to be large enough to achieve positive supremum for all instances, where $P_0 \neq P_1$. The choice of $\mathcal{F}$ leads to crucial distinctions between IPMs \cite{sriperumbudur2012empirical}, a few examples of which are as follows:

\begin{itemize}
    \item If we set $\mathcal{F}$ to be all functions over $\mathcal{X}$ bounded by $1$, \eqref{eq:IPM} recovers the \emph{Total Variation Distance} (TVD).
    \item If we set $\mathcal{F}$ to be all $1$-Lipschitz functions over $\mathcal{X}$, \eqref{eq:IPM} recovers the \emph{Wasserstein Distance} (WD) \cite{arjovsky2017wasserstein}.
    \item If we set $\mathcal{F}$ to be the unit ball of a Reproducing Kernel Hilbert Space (RKHS) \cite{aronszajn1950theory}, \eqref{eq:IPM} recovers the \emph{Maximum Mean Discrepancy} (MMD) \cite{gretton2012kernel}.
\end{itemize}

Of these three examples, we focus on WD and MMD because of their sensitivity to the topology of the distributions $P_0$, $P_1$ \cite{li2017mmd}.\footnote{TVD is not affected by the closeness of the supports of $P_0$ and $P_1$ when they are disjoint. This makes TVD a difficult objective to be utilized with gradient based techniques, because there will not be suitable descent directions in areas where the supports do not overlap.}

\subsubsection{Maximum Mean Discrepancy (MMD)} \label{sec:mmd}

It is convenient to start with MMD due to the closed form expressions of its estimates. Since the authors in \cite{gretton2012kernel} found that the biased estimate of (squared) MMD works much better than its unbiased alternative, we utilize the biased statistic in \cite{gretton2012kernel} while defining the MMD objective. Assuming we have $N$ samples $\{(\bz_i,s_i)\}_{i=1}^{N}$ with $s_i\in \{0,1\}$ and $N_j$ samples for which $s_i=j$, we can define the MMD statistic as
\begin{multline}
    \text{MMD} = \Bigg( \frac{1}{N_0^2}\sum_{s_i=0}\sum_{s_j=0} k(\bz_i,\bz_j) + \frac{1}{N_1^2}\sum_{s_i=1}\sum_{s_j=1} k(\bz_i,\bz_j) \\- \frac{2}{N_0 N_1}\sum_{s_i=0}\sum_{s_j=1} k(\bz_i,\bz_j)\Bigg)^{\nicefrac{1}{2}} \text{.}
    \label{eq:MMD_sq}
\end{multline}

The MMD statistic \eqref{eq:MMD_sq} is simply a closed form expression of the IPM \eqref{eq:IPM} when $\mathcal{F}$ is the unit ball of an RKHS and $P_0$, $P_1$ are empirical distributions of the data. To see this, let us first express the IPM based on our sample,
\begin{equation}
    \text{MMD}_{smpl} = \underset{f \colon \lnorm f \rnorm_{\mathcal{H}} \leq 1}{\sup} \left\{ \frac{1}{N_0} \sum_{s_i=0} f(\bz_i) - \frac{1}{N_1} \sum_{s_i=1} f(\bz_i) \right\} \text{,}
    \label{eq:mmd_empirical}
\end{equation}
with $\mathcal{H}$ denoting an RKHS. Since the data is drawn from a compact set $\mathcal{X}$, the mean emdeddings $\bmu_0 \coloneqq \frac{1}{N_0}\sum_{s_i=0} k(\bz_i,\cdot)$ and $\bmu_1 \coloneqq \frac{1}{N_1}\sum_{s_i=1} k(\bz_i,\cdot)$ will be inside the RKHS. Then, by Riesz Representer Theorem \cite{reed1980methods}, we can rewrite \eqref{eq:mmd_empirical} as
\begin{align}
    &\text{MMD}_{smpl} &\\ 
    &= \underset{f \colon \lnorm f \rnorm_{\mathcal{H}} \leq 1}{\sup} \left\langle \frac{1}{N_0}\sum_{s_i=0} k(\bz_i,\cdot)-\frac{1}{N_1}\sum_{s_i=1} k(\bz_i,\cdot), f \right\rangle_{\mathcal{H}} &\\ 
    &= \lnorm \frac{1}{N_0}\sum_{s_i=0} k(\bz_i,\cdot)-\frac{1}{N_1}\sum_{s_i=1} k(\bz_i,\cdot) \rnorm_{\mathcal{H}} &\\
    &= \text{MMD}  \text{,}&
\end{align}
where MMD is as defined in \eqref{eq:MMD_sq} and the last equality is due to the reproducing property $\langle k(\bz_i,\cdot), k(\bz_j,\cdot) \rangle_{\mathcal{H}}=k(\bz_i,\bz_j)$.

The MMD defines a proper metric on a compact set $\mathcal{X}$ when the RKHS $\mathcal{H}$ is \textit{universal}. An RKHS is universal when $k(\cdot,\cdot)$ is continuous and $\mathcal{H}$ is dense in the set of all continuous functions \cite{gretton2012kernel}. Clearly, \textit{universal approximators} like Gaussian and Laplacian are also universal kernels, since their RKHSs are dense in the set of all measurable functions. The fact that universality leads to a proper metric on a compact set is a consequence of $\text{MMD}(P_0,P_1)=0$ being equivalent to $P_0$ and $P_1$ having all their moments equal. Note that matching all their moments is sufficient for equalizing distributions, provided the distributions have supports over compact sets. 

As a sidenote we would like to add that, while weaker kernels like polynomials do not define proper metrics, their MMD still has an extremely useful interpretation. For a $4^{th}$ order polynomial kernel $k(\bz_i,\bz_j)=\left(1+\gamma\left(\bz_i^\top\bz_j\right)\right)^4$, for instance, $\text{MMD}(P_0,P_1)=0$ implies $P_0$, $P_1$ have matching mean, variance, skew and kurtosis \cite{gretton2012kernel,sriperumbudur2012empirical}. Therefore, MMD can be used to remove correlations up to a certain order, if such processing is known to be sufficient.

\subsubsection{Wasserstein Discriminator Network (WDN)} \label{sec:wdn}

WD is a measure often used in Generative Adversarial Network (GAN) training \cite{arjovsky2017wasserstein,gulrajani2017improved,shen2018wasserstein}. While WD does not lead to suitable closed form estimates from data, the maximization in \eqref{eq:IPM} can be performed over a parametric family of functions $\mathcal{F}\coloneqq \{f(\cdot;\btheta)\colon \btheta \in \Theta\}$. Accordingly, by selecting an expressive neural network architecture and appropriate regularization, we could approximate the maximization over all $C$-Lipschitz functions for some constant $C$ \cite{arjovsky2017wasserstein,shen2018wasserstein,wei2018improving}.\footnote{Note that the Lipschitz constant simply scales the WD, hence its exact value is not important.}

The network in question maximizes the linear loss over a class of functions $\mathcal{F}$ defined by its architecture and its regularization. Namely, for a binary valued private attribute $S$, it tries to assign positive values to samples with $S=0$ and negative values to samples with $S=1$. Since this network effectively tries to discriminate the privacy class from data, we name it the \textit{privacy discriminator}. For the particular privacy discriminator that approximates the Wasserstein Distance, we use the name \textit{Wasserstein Discriminator Network} (WDN).

Assuming we have $N$ samples $\{(\bz_i,s_i)\}_{i=1}^{N}$ with $s_i\in \{0,1\}$ and $N_j$ samples for which $s_i=j$, the first part of the WDN loss is
\begin{equation}
    L_D(\bZ,\bs;\btheta_D) = \frac{1}{N_1}\sum_{s_i=1} \phi_D(\bz_i;\btheta_D)-\frac{1}{N_0}\sum_{s_i=0} \phi_D(\bz_i;\btheta_D) \text{,}
    \label{eq:wd_linear}
\end{equation}
where $\bZ=[\bz_1 \ldots \bz_N]$ is the processed data matrix, $\bs=[s_1 \ldots s_N]^\top$ is the privacy label vector and $\btheta_D$ are the parameters of the WDN network. Note that the parameters $\btheta_D$ minimizing the loss come after the semi-colon.

To ensure that the WDN obeys the Lipschitz constraint, we also need to apply some regularization. The first networks approximating WD maintained Lipschitz functions via weight clipping \cite{arjovsky2017wasserstein}, and later works have introduced gradient penalties \cite{shen2018wasserstein}, as well as other terms that ensure the network is consistent with the Lipschitz constraint on and around the input samples $\{\bz_i\}_{i=1}^{N}$ \cite{wei2018improving}. We utilize the gradient penalty introduced in \cite{shen2018wasserstein}, which can be written as
\begin{equation}
    L_R(\bZ;\btheta_D) = \frac{1}{N}\sum_{i=1}^{N} \lp \lnorm \nabla_{\bz_i} \phi_D(\bz_i;\btheta_D) \rnorm_2 - 1  \rp^2 \text{,}
    \label{eq:wd_lipschitz}
\end{equation}
where $\nabla_{\bz} \phi_D(\bz;\btheta_D)$ denotes the gradient of the WDN function $\phi_D(\cdot;\btheta_D)$ with respect to its input. Notice that \eqref{eq:wd_lipschitz} forces the norm of the gradient to be close to 1 instead of being smaller than 1, which is actually what being $1$-Lipschitz is equivalent to for differentiable functions. This is due to the observation in the original work \cite{shen2018wasserstein}, which finds the two-sided penalty to work slightly better. It is argued that the extra penalty likely does not lead to a significant constraint on the discriminator.

It is well-known that such penalties only force the function $\phi_D(\cdot;\btheta_D)$ to be Lipschitz on the data samples $\{\bz_i\}_{i=1}^{N}$ and not in general. This is tolerable given we are only dealing with empirical distributions, and we do not apply the discriminator function across the full support of the underlying distributions. It is worth keeping in mind, however, that we can extend the Lipschitz constraint to the vicinity of the samples by adding noise to $\bz_i$ in \eqref{eq:wd_lipschitz}.

With the linear discriminator loss \eqref{eq:wd_linear} and the Lipschitz regularizer \eqref{eq:wd_lipschitz}, the overall loss of the WDN is given by
\begin{equation}
    L_{Disc}(\bZ,\bs;\btheta_D) = L_D(\bZ,\bs;\btheta_D) + \lambda L_R(\bZ;\btheta_D),
    \label{eq:wdn_loss}
\end{equation}
where $\lambda$ is the regularization parameter. Our privacy enhancing feature maps then try to drive the discriminator loss to be as high as possible to ensure the private information cannot be inferred successfully. Therefore, the (adversarial) privacy loss of our data desensitizing network $\phi_P(\cdot;\btheta_P)$ is given by
\begin{multline}
    L_P(\bX,\bs,\btheta_D;\btheta_P) = -\frac{1}{N_1}\sum_{s_i=1} \phi_D(\phi_P(\bx_i;\btheta_P);\btheta_D)\\+\frac{1}{N_0}\sum_{s_i=0} \phi_D(\phi_P(\bx_i;\btheta_P);\btheta_D)\text{.}
    \label{eq:wdn_priv_loss}
\end{multline}
Note that the WDN parameters $\btheta_D$ come before the semi-colon, hence, they are treated as constants here, with $\btheta_P$ being the optimization parameters. The parameters $\btheta_P,\btheta_D$ are, thus, optimized jointly towards the two opposing goals: Hiding private information and inferring private information, respectively.

\subsubsection{Using IPMs with $L$-ary Private Attributes} \label{sec:multi-class}

We employ a \textit{One-vs-Rest} approach to generalize the binary MMD and WD objectives to $L$-ary objectives. Assume that $S \in \{0,1,\ldots,L-1\}$, define $\pi_i=\mathbb{P}[S=i]$, $P_i=P_{Z|S=i}$ and $Q_i=P_{Z|S \neq i}$. Let $Z_i \sim P_i$ and $\bar{Z}_i \sim Q_j$. A combined objective can be written as the weighted linear combination of IPMs
\begin{equation}
    \sum_{i=0}^{L-1} \pi_i \underset{f \in \mathcal{F}}{\sup} \left\{ \bbE\left[f(Z_i)\right]-\bbE\left[f(\bar{Z}_i)\right] \right\} \text{.}
    \label{eq:multi_IPM}
\end{equation}
Notice that if $S$ is binary, \eqref{eq:multi_IPM} reduces to \eqref{eq:IPM} due to the function classes we consider being closed under additive inverses. 

We apply this generalization to the MMD objective \eqref{eq:MMD_sq} to yield a multi-class version. Assuming we have $N$ samples $\{(\bz_i,s_i)\}_{i=1}^{N}$ with $s_i\in \{0,1,\ldots,L-1\}$ and $N_l$ samples for which $s_i=l$,
\begin{multline}
    \sum_{l=0}^{L-1} \frac{N_l}{N} \Bigg( \frac{1}{N_l^2}\sum_{s_i=l}\sum_{s_j=l} k(\bz_i,\bz_j) + \frac{1}{\overline{N}_l^2}\sum_{s_i \neq l}\sum_{s_j \neq l} k(\bz_i,\bz_j) \\- \frac{2}{N_l \overline{N}_l}\sum_{s_i=l}\sum_{s_j \neq l} k(\bz_i,\bz_j) \Bigg)^{\nicefrac{1}{2}}  ,
    \label{eq:multi_MMD}
\end{multline}
where $\overline{N}_l=N-N_l$ is the size of the complement class. Once again, this yields the binary MMD \eqref{eq:MMD_sq}, if $s_i \in \{0,1\}$.

For the Wasserstein Discriminator Network (WDN), we generalize the discriminator loss \eqref{eq:wd_linear} and the Lipschitz regularizer \eqref{eq:wd_lipschitz} separately. Our corresponding discriminator loss is 
\begin{multline}
    L_D(\bZ,\bs;\btheta_D) = \sum_{l=0}^{L-1} \frac{N_l}{N} \Bigg( \frac{1}{N_l}\sum_{s_i=l} {\phi_D}_l(\bz_i;\btheta_D) \\- \frac{1}{\overline{N}_l}\sum_{s_i \neq l} {\phi_D}_l(\bz_i;\btheta_D) \Bigg)\text{,}
    \label{eq:wd_multi_linear}
\end{multline}
where we denote by ${\phi_D}_l$ the $(l+1)^{th}$ output of the WDN. The discriminator thus has $L$ outputs, each meant to distinguish one privacy class from the rest. We should normally train a separate network for each class (or $L-1$ networks to avoid redundancy) for \eqref{eq:wd_multi_linear} to be consistent with \eqref{eq:multi_IPM}. Having a shared network is much more computationally efficient, however, especially in instances where the number of privacy classes is large. Therefore, we make a compromise to utilize the shared structure between the $L$ discriminator tasks.

For the Lipschitz regularizer, we elected to apply the gradient penalty to the output corresponding to the privacy class of a sample. The reason for this is that, if we apply $L$ gradient penalties per sample, the amount of memory usage required for back-propagation can grow very large in instances with large $L$. To avoid significant constraints on the usable batch sizes, we thus apply one gradient penalty per sample. Since much of the WDN structure is shared between privacy classes (with only difference being the final linear mapping), this type of regularization still suffices to achieve a Lipschitz constant on all the network outputs. The resulting regularizer is
\begin{equation}
    L_R(\bZ,\bs;\btheta_D) = \frac{1}{N}\sum_{i=1}^{N} \lp \lnorm \nabla_{\bz_i} {\phi_D}_{s_i}(\bz_i;\btheta_D) \rnorm_2 - 1  \rp^2 \text{.}
    \label{eq:wd_multi_lipschitz}
\end{equation}

The overall privacy discriminator loss is once again the sum of the discriminator loss and the Lipschitz regularizer as in \eqref{eq:wdn_loss}. The privacy objective of the desensitizing network $\phi_P(\cdot;\btheta_P)$ is also the inverse of the discriminator loss as in \eqref{eq:wdn_priv_loss},
\begin{multline}
     L_P(\bX,\bs,\btheta_D;\btheta_P) \\=  \sum_{l=0}^{L-1} \frac{N_l}{N} \Bigg( -\frac{1}{N_l}\sum_{s_i=l} {\phi_D}_l(\phi_P(\bx_i;\btheta_P);\btheta_D) \\+ \frac{1}{\overline{N}_l}\sum_{s_i \neq l} {\phi_D}_l(\phi_P(\bx_i;\btheta_P);\btheta_D) \Bigg) \text{.}
     \label{eq:wd_multi_priv}
\end{multline}

\subsection{Least Squares Based Criteria} \label{sec:ls_criteria}

We established MMD and WD as feasible privacy objectives, if the private variables are discrete. Kernel Ridge Regression (KRR) \cite{al2020scalable} and the Least Squares Generative Adversarial Networks \cite{mao2017least} provide additional objectives suitable for the more general case, where the private variables can be continuous. Hence, we will present the Kernel Discriminant Information (KDI) and the Least Squares Discriminator Networks (LSDNs) as additional tools for enhancing privacy. 

\subsubsection{Kernel Discriminant Information (KDI)} \label{sec:kdi}

We start covering the least squares-based criteria with the Kernel Discriminant Information (KDI) \cite{kung2017compressive}, because it provides us with a closed form privacy objective similar to MMD. Let us denote the privacy label matrix by $\bP$ and consider a KRR predictor as the privacy discriminator. We can then express the minimum loss of the privacy discriminator (MLPD) as
\begin{equation}
 \text{MLPD} = \underset{\bW,\bb}{\text{min}} \left\| \bPhi^\top\bW+\rone\bb^\top-\bP \right\|_F^2+\rho\left\|\bW\right\|_F^2 \text{,}
\label{eq:PHSRR}
\end{equation}
where $\bPhi \coloneqq \left[\phi_k(\bz_1) \ldots \phi_k(\bz_N) \right]$ is a matrix containing RKHS mappings of the processed data samples $\{\bz_i\}_{i=1}^{N}$ such that $\phi_k^\top(\bz_i) \phi_k(\bz_j)=k(\bz_i,\bz_j)$.

Setting the gradients equal to zero yields the optimal bias vector $\bb^*=N^{-1}\lp \bP^\top \rone-\bW^\top \bPhi \rone \rp$ and weight matrix $\bW^* = \lp \Sbar+\rho\bI \rp^{-1}\bar{\bPhi}\bar{\bP}$, with $\Sbar=\bar{\bPhi}\bar{\bPhi}^\top$, $\bar{\bPhi}=\bPhi \bC$, $\bar{\bP} = \bC \bP$ and $\bC = \bI-\frac{1}{N}\rone\rone^\top$. Notice that $\bar{\bPhi}\bar{\bP}=\bPhi \bC \bC \bP=\bPhi \bC \bP=\bar{\bPhi}\bP$. Upon plugging in the optimal solution to the minimization in \eqref{eq:PHSRR}, we can express the MLPD as
\begin{equation}
\text{MLPD} = -\trace\lp \lp \Sbar+\rho \bI \rp^{-1} \Sb \rp + \lnorm\bar{\bP}\rnorm_F^2 \text{,}
\label{eq:pmrlse_di}
\end{equation}
where $\Sb=\bar{\bPhi}\bP \bP^{\top}\bar{\bPhi}^\top$. $\Sb$ is the well known between-class scatter matrix when $\bP$ is a class indicator matrix with each column scaled to be unit norm. However, this definition naturally encompasses the regression setting with arbitrary $\bP$. Ignoring the constant term, we see that the minimum loss of the privacy discriminator can be maximized by minimizing the quantity we refer to as \emph{the Discriminant Information} (DI), which is $\trace\lp \lp \Sbar+\rho \bI \rp^{-1} \Sb \rp$ \cite{al2020scalable}.

For our purposes, we shall define a kernelized equivalent of DI (i.e., KDI), which does not rely on explicit RKHS mappings of the data. For this, we first express $\bU \bSigma \bV^\top = \bar{\bPhi}$ as the compact SVD of the matrix $\bar{\bPhi}$. Noting that $\Sbar=\bar{\bPhi}\bar{\bPhi}^\top=\bU \bSigma^2 \bU^\top$, $\Kbar=\bar{\bPhi}^\top \bar{\bPhi} = \bV \bSigma^2 \bV^\top$, $\bU^\top \bU = \bI = \bV^\top \bV$ and using the cyclical property of the trace, we get
\begin{align}
    & \text{KDI} \equiv \text{DI} \nonumber  \\  
    & =  \trace\lp \lp \Sbar+\rho \bI \rp^{-1} \Sb \rp  \\
    & = \trace \lp \bSigma \bU^\top \bU \lp \bSigma^2 +\rho \bI \rp^{-1} \bU^\top \bU \bSigma \bV^\top \bP \bP^\top \bV \rp\\
    & = \trace \lp \bSigma^2 \bV^\top \bV \lp \bSigma^4 +\rho \bSigma^2 \rp^{-1} \bV^\top \bV \bSigma^2 \bV^\top \bP \bP^\top \bV \rp\\
    & = \trace\lp \lp \Kbar^2+\rho \Kbar \rp^{+} \Kb \rp \text{,} \label{eq:KDI}
\end{align}
where $\Kb=\Kbar \bP \bP^\top \Kbar$. We use the expression of KDI given by \eqref{eq:KDI}, which allows us to plug in the centered kernel matrix as a function of the processed data matrix: $\Kbar=\bC k(\bZ,\bZ) \bC$.

The ridge regularizer $\rho \|\bW\|_F^2$ in \eqref{eq:PHSRR} plays the $l_2$ regularizer role in the Reproducing Kernel Hilbert Space (RKHS), hence, it has the effect of constraining the function class into some $l_2$ ball \cite{scholkopf2001generalized}. For this reason, KDI measures the minimum least-squares error achieved in some $l_2$ ball of the RKHS, similar to how MMD measures the minimum linear loss achieved in the unit $l_2$ ball. 
 
 In the binary classification setting, the main practical difference between MMD and KDI is that MMD directly measures the Euclidean distance between mean embeddings in an RKHS, whereas KDI measures the Euclidean distance after \textit{whitening}. That is, KDI multiplies the mean embeddings with the square root of the inverse of the sample covariance matrix (with a ridge regularizer added). To see this, consider the binary case where $\bP \coloneqq \mathbf{p}$ is a vector containing the class labels $0$ and $1$. Then, we have
 \begin{align}
     \trace(\Sb) &= \mathbf{p}^\top \Phibar \Phibar^\top \mathbf{p} \\ 
     &= N_1^2 \lnorm \bmu_1-\bmu \rnorm^2  \\ 
     &= \frac{N_1^2 N_0^2}{N^2} \lnorm \bmu_1 -\bmu_0 \rnorm^2 \text{,}
 \end{align}
 where $\bmu \coloneqq \frac{1}{N} \sum_{i=1}^{N} \phi_k(\bz_i)$ denotes the overall mean; $\bmu_j\coloneqq \frac{1}{N_j} \sum_{s_i=j} \phi_k(\bz_i)$, $N_j$ denote the mean and number of samples of the privacy class $j$, respectively. Except for the constant factor $\delta=(\nicefrac{N_1N_0}{N})^2$, $\trace(\Sb)$ then yields the square of the $\text{MMD}$ statistic in \eqref{eq:MMD_sq} after utilizing the kernel trick. For many commonly used kernels, this relationship allows us to bound\footnote{These bounds on $\text{KDI}$ are derived in the binary classification setting. We omit their generalization to the multi-class setting here, though it can be obtained by expressing multi-class KDI as a sum of binary KDI.} the KDI above and below in terms of $\text{MMD}^2$. For the Gaussian kernel, for example, the following result is obtained.
 \begin{align}
     \frac{\delta}{N+\rho}\text{MMD}^2 &=  \frac{\delta}{N+\rho} \lnorm \bmu_1 -\bmu_0 \rnorm^2 & \label{eq:mmd_bound1}\\
     & \leq \lambda_{min} \lp \lp \Sbar +\rho \bI \rp^{-1} \rp \trace \lp \Sb \rp & \label{eq:mmd_bound2}\\
     & \leq \trace\lp \lp \Sbar+\rho \bI \rp^{-1} \Sb \rp & \label{eq:mmd_bound3}\\
     & \leq \lambda_{max} \lp \lp \Sbar +\rho \bI \rp^{-1} \rp \trace \lp \Sb \rp & \label{eq:mmd_bound4}\\
     & \leq \frac{\delta}{\rho} \lnorm \bmu_1 -\bmu_0 \rnorm^2 = \frac{\delta}{\rho}\text{MMD}^2 \text{,} \label{eq:mmd_bound5}
 \end{align}
 where for the second and third inequalities, we used the fact that the eigenvalues of $\Sbar+\rho \bI$ are bounded above and below by $N+\rho$ and $\rho$, respectively, which also bounds the inner product between symmetric positive semi-definite matrices \cite{golub2013matrix}. The bound on the eigenvalues is a consequence of the Gaussian kernel matrix having unit diagonals and the fact that $\Sbar$, $\Kbar$ have the same non-zero eigenvalues.\footnote{Unit diagonals ensure that $\trace(\bK)=N$, since the trace is the sum of eigenvalues, the maximum eigenvalue has to be smaller than $N$. Additionally, $\lambda_{max}(\Sbar) = \lambda_{max}(\Kbar) = \lambda_{max}(\bC \bK \bC) \leq \lambda_{max} (\bK)$, since $\bC$ is another symmetric positive semi-definite matrix with $\lambda_{max}(\bC) = 1$.} Similar bounds can be established for all other kernels when they are applied to compact data domains, as the maximum eigenvalues will always be bounded in such instances. For all the RBF kernels satisfying $k(\bx,\bx)=1$, the same result applies,
 \begin{equation}
     \frac{\delta}{{N+\rho}}\text{MMD}^2 \leq \text{KDI} \leq \frac{\delta}{{\rho}}\text{MMD}^2 \text.
     \label{eq:kdi_bound}
 \end{equation}
 
While MMD and KDI will have different performances on finite samples, they are both consistent statistics for testing the equality of distributions when used with universal kernels \cite{gretton2012kernel,eric2008testing}. For this reason, and the bound we established in \eqref{eq:kdi_bound}, KDI is a suitable alternative to MMD, which also generalizes to continuous variables.

\subsubsection{Least Squares Discriminator Network (LSDN)}

Another privacy discriminator we consider is the \emph{Least Squares Discriminator Network} (LSDN), which is a neural network minimizing the squared error of its predictions \cite{mao2017least}. The objective of LSDN can be written as
\begin{equation}
    L_D (\bZ,\bP;\btheta_D) = \frac{1}{N}\sum_{i=1}^{N} \lnorm \phi_D(\bz_i;\btheta_D)-\bs_i \rnorm^2 \text{.}
    \label{eq:lsdn_loss}
\end{equation}

Differently from the WDN, the privacy objective of the data desensitizing network is not given simply by the inverse of the discriminator loss. Instead, the data desensitization explicitly tries to make all predictions of the privacy discriminator the same as the mean prediction, that is, the best prediction LSDN could make if $\bZ=\phi_P(\bX;\btheta_P)$ contained no information on the privacy labels $\bP$. Therefore, the (adversarial) privacy loss minimized by $\btheta_P$ is given by
\begin{equation}
    L_P (\bX,\bP,\btheta_D;\btheta_P) = \frac{1}{N}\sum_{i=1}^{N} \lnorm \phi_D(\phi_P(\bx_i;\btheta_P);\btheta_D)-\bmu \rnorm^2 \text{,}
    \label{eq:lsdn_priv_loss}
\end{equation}
where $\bmu = \frac{1}{N}\sum_{i=1}^{N}\bs_i$. This definition of the privacy loss leads to a functionality similar to that of KDI. KDI \eqref{eq:KDI} being $0$ implies the best KRR predictor always predicts the target mean $\bmu$, whereas, the privacy loss \eqref{eq:lsdn_priv_loss} being $0$ implies the LSDN $\phi_D(\cdot;\theta_D)$ always predicts the target mean.

In the binary classification setting, where $s_i \in \{0,1\}$, minimizing \eqref{eq:lsdn_priv_loss} against a discriminator minimizing \eqref{eq:lsdn_loss} was shown, in effect, to be a minimization of Pearson's $\chi^2$ divergence between distributions \cite{mao2017least}. Our definition of the losses allows privacy enhancing feature maps to be optimized based on continuous-valued private attributes as well.

\section{Methodology}\label{sec:methodology}

\subsection{Network Architectures}
\begin{figure}[t]
  \begin{minipage}[b]{\linewidth}
   \centering
   \centerline{\includegraphics[trim = 150 40 150 40,clip,width=\linewidth]{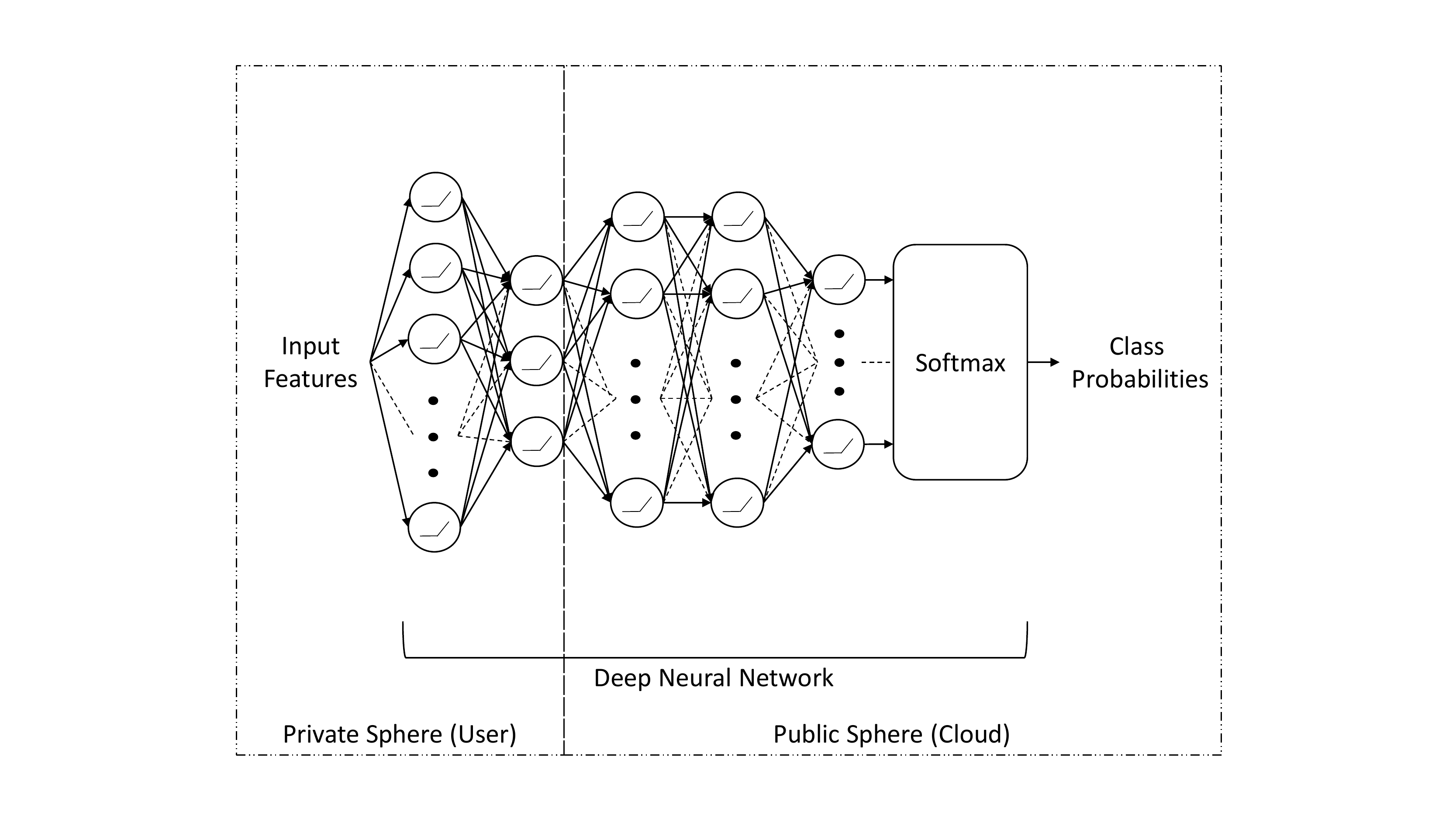}}
  \end{minipage}
  \caption[Illustration of Private/Public Sphere Separation of a Network]{The schematic of a neural network classifier separated between private and public spheres.}
  \label{fig:StdNeuralNet}
\end{figure}

In Sections \ref{sec:ipms} and \ref{sec:ls_criteria}, we have introduced multiple privacy objectives that can help remove dependencies between released data and private variables. The next step is to combine these training objectives with feature maps and utility objectives so that we can ensure the data serves the intended utility goal after being cleaned from unwanted dependencies.

To produce a system that meets the utility goals of users without revealing their private information, the data desensitization has to be performed in the \emph{private sphere}. After the desensitization process, predictive models can be applied to the data to infer information that is desirable to the user. To seamlessly incorporate the optimization of the public and private sphere models, it is natural to consider the two parts as a single feed-forward network, which are separated only by where the computations are performed. A simple structure of this sort is displayed in Figure \ref{fig:StdNeuralNet}. 
The private sphere ends with a narrow, funneling layer, which outputs a low-dimensional representation of the original data. This low-dimensional output also constitutes the input of the public sphere, which does not get access to the original data. 

The privacy objectives described in Sections \ref{sec:ipms} and \ref{sec:ls_criteria} can be applied to the output of the private sphere to ensure that the low-dimensional representations that are sent out reveal minimal private information. The utility objectives, on the other hand, are best applied to the outputs of the public sphere, since the end goal is to get the most accurate predictions out of the entire network.

For dense neural networks (DNN), no special structure needs to be applied other than a narrow layer at the end of the private sphere. Hence, the mappings learned in the public and private spheres are highly flexible. For our experiments, we apply a softmax layer to the outputs of the public sphere for classification goals, and we use the Rectified Linear Units (ReLU) \cite{nair2010rectified}, i.e., $f(x)=\max(0,x)$, as activations for all hidden layers, including the narrow, funneling layer. This choice is partially motivated by the fact that we found non-invertible activation functions to improve the privacy performances of the resulting feature maps.

For convolutional neural network (CNN) architectures, an important characteristic to keep in mind is that each feature is typically a function of a small subset of input pixels. However, for the effective removal of sensitive information, output features generally need to be global functions of the input features. For CNNs to produce high level representations that are global functions of the inputs, many layers of convolutions are typically needed. This in turn can result in a huge computational overhead on the user side. Hence, to achieve relatively shallow convolutional feature mappings that produce global functions of the input pixels, we try to incorporate dense layers into CNNs without removing spacial correlations between pixels. Subspace projections enable us to perform such dense mappings in image domains \cite{yu2006learning,chanyaswad2016discriminant}.

We can achieve subspace projections between convolutional layers by inserting a mapping
\begin{equation}
 \begin{split}
 \phi(\bx) = \bW^\top h(\bx)
 \end{split}
 \quad \text{and} \quad
 \begin{split}
  \widetilde{h}(\bx) = \bW\phi(\bx)\text{,}
 \end{split}
\end{equation}
where $\bW$ has orthonormal columns ($\bW^\top \bW = \bI$), $h(\bx)$ denotes a flattened hidden layer and $\widetilde{h}(\bx)$ denotes its reconstruction from $\phi(\bx)$. The addition of orthonormal projections provides a cheap way of reconstructing images from their dense mappings based on the squared error criterion, and we achieve orthonormality in the projection matrix $\bW$ by adding the following penalty to our training objective
\begin{equation}
    L_O = \lnorm \bW^\top \bW - \bI \rnorm_F^2
    \label{eq:orthonormality}
\end{equation}

In our experiments, a penalty factor of $10$ sufficed to obtain nearly orthonormal projection matrices with the penalty \eqref{eq:orthonormality} becoming less than $10^{-4}$. To give the network more freedom in choosing active projection directions, we also add ReLU activations and bias terms
\begin{equation}
 \begin{split}
 \phi(\bx) = ReLU(\bW^\top h(\bx)+\bb)
 \end{split}
 \quad \text{and} \quad
 \begin{split}
  \widetilde{h}(\bx) = \bW\phi(\bx)\text{,}
 \end{split}
 \label{eq:cnn_funnel}
\end{equation}
where $\phi(\bx)$ denotes the output of the private sphere. With this mapping, projection directions whose component values fall below a certain threshold get discarded, hence, the reconstructions can be based on a smaller number of projection directions than the number of columns of $\bW$. We found that this modification improves the privacy performances of CNNs within the private sphere without hindering the ultimate utility performance of the system as a whole. We do not use these orthonormal projection layers in dense neural networks, since we are not worried about maintaining spacial correlations between dense layers. 

\subsection{Network Objectives for Utility and Privacy}\label{subsec:Network Obj}

\begin{figure*}[t]
  \begin{minipage}[b]{\linewidth}
   \centering
   \centerline{\includegraphics[trim = 70 60 70 60,clip,width=0.8\linewidth]{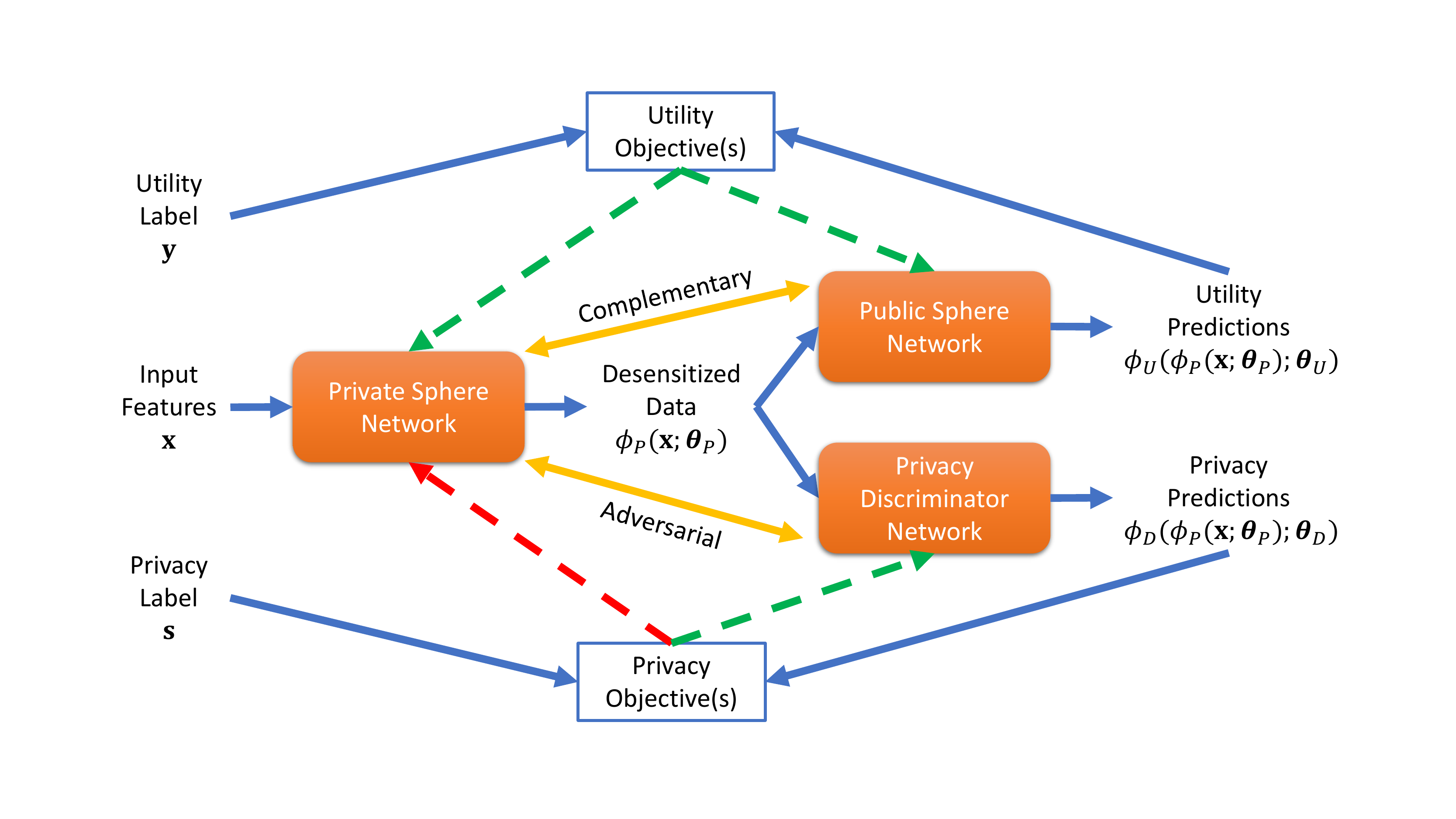}}
  \end{minipage}
  \caption[Structure of the End-to-End Optimization System]{The structure of the optimized system. The private sphere network feeds the public sphere network and the privacy discriminator, whose outputs are used to compute the utility and privacy objectives, respectively. A green dashed line indicates the network serves the objective(s), while a red dashed line indicates the network hinders the objective(s).}
  \label{fig:Structure}
\end{figure*}

In this section, we summarize the training objectives of predictive neural network models, which are split between a private sphere and a public sphere. For a general treatment, we consider the tuple $(\bX,\bY,\bP)$ as our training dataset, which consists of an $N$-columned data matrix $\bX$, and $N$-rowed utility and privacy label matrices $\bY$ and $\bP$, respectively. Accordingly, $\bx_i$, $\by_i$ and $\bs_i$ refer to the input features, utility label and privacy label for the $i^{th}$ sample, respectively. To denote the mappings performed by the \emph{private sphere network}, the \emph{public sphere network} and the \emph{privacy discriminator network}, we use $\phi_P(\bx;\btheta_P)$, $\phi_U(\phi_P(\bx;\btheta_P);\btheta_U)$ and $\phi_D(\phi_P(\bx;\btheta_P);\btheta_D)$, respectively, where $\btheta_P$, $\btheta_U$ and $\btheta_D$ are the parameters of these respective networks. We also remind that we use $\phi(\bZ;\btheta) \coloneqq [\phi(\bz_1;\btheta) \ldots \phi(\bz_N;\btheta)]$ to represent an entire data matrix $\bZ$ after being processed by a network $\phi(\cdot;\btheta)$.

The structure of our optimized system is summarized in Figure \ref{fig:Structure}. The public and private sphere networks both serve the utility prediction goal, hence, they are complementary in nature. The privacy discriminator, on the other hand, serves the privacy prediction goal, which the private sphere network tries to hinder, thus, these networks are adversarial in nature. Note that the privacy discriminator only exists in settings where we use the adversarial WDN \eqref{eq:wd_multi_priv} and LSDN \eqref{eq:lsdn_priv_loss} losses as our privacy objectives. The outputs of the (kernel based) privacy discriminators are already incorporated into the MMD \eqref{eq:multi_MMD} and KDI \eqref{eq:KDI} objectives, hence, no explicit privacy discriminator is needed for optimizing them. Below, we go over the utility and privacy objectives and the networks optimizing them.

\subsubsection{The Utility Objective for the Private and Public Spheres}
We use the traditional utility objectives for training neural network predictors, namely, the Cross-Entropy (CE) loss for classification tasks. With the utility label $\by_i$ being a vector containing class probabilities, the utility loss function is
\begin{equation}
    L_U(\bX,\bY;\btheta_P,\btheta_U) = -\frac{1}{N}\sum_{i=1}^{N} \by_i^\top \log(\phi_U(\phi_P(\bx_i;\btheta_P);\btheta_U)) \text{,}
    \label{eq:util_loss}
\end{equation}
where $\log(\cdot)$ denotes the element-wise logarithm. This utility objective is always applied when optimizing $\btheta_P,\btheta_U$, regardless of the type of privacy objective and network architecture used. 

\subsubsection{The Privacy Objectives for the Private Sphere}\label{subsubsec:Priv Obj}

\begin{table}[t]
  \caption[Summary of Privacy Objectives]{Summary of the privacy objectives.}
  %\small
  \centering
  \begin{tabularx}{\linewidth}{m{0.6cm} C{1.5cm} C{1.8cm} Y}
  \toprule
  Obj. & Discriminator & Variable Type & Objective Format \\
  \midrule
  MMD & Kernel Net. & Discrete & Closed Form Statistic \eqref{eq:MMD_obj} \\
  KDI & Kernel Net. & Discrete / Cont. & Closed Form Statistic \eqref{eq:KDI_obj}  \\
  WDN & Neural Net. & Discrete & Adversarial Loss \eqref{eq:WDN_obj}  \\
  LSDN & Neural Net. & Discrete / Cont. & Adversarial Loss \eqref{eq:LSDN_obj} \\
   \bottomrule
  \end{tabularx}
  %\normalsize
  \label{tab:priv_objectives}
\end{table}

Below are the privacy objectives we utilize for optimizing $\btheta_P$. To give readers a simplified perspective on these, we summarize their key properties in Table \ref{tab:priv_objectives}.

\textbf{MMD}: We use the privacy objective in \eqref{eq:multi_MMD}, but express it more succintly in matrix form. Let the privacy label matrix $\bP$ contain the one-hot encodings of privacy classes in its rows. Let $\bp_l$ denote the $l^{th}$ column of the $N\times L$ matrix $\bP$, and $\overline{\bp}_l=\rone-\bp_l$, then the privacy loss function is
\begin{multline}
    L_P(\bX,\bP;\btheta_P) \\= \sum_{l=1}^{L} \frac{N_l}{N} \Bigg( \frac{1}{N_l^2} \bp_l^\top \bK(\bX;\btheta_P) \bp_l + \frac{1}{\overline{N}_l^2}\overline{\bp}_l^\top \bK(\bX;\btheta_P) \overline{\bp}_l \\- \frac{2}{N_l \overline{N}_l} \bp_l^\top \bK(\bX;\btheta_P) \overline{\bp}_l \Bigg)^{\nicefrac{1}{2}}  \text{,}
    \label{eq:MMD_obj}
\end{multline}
where $N_l=\rone^\top \bp_l$,  $\overline{N}_l=\rone^\top \overline{\bp}_l$, and $\bK(\bX;\btheta_P)=k(\phi_P(\bX;\btheta_P),\phi_P(\bX;\btheta_P))$ is the $N\times N$ kernel matrix obtained by applying a kernel function $k(\cdot,\cdot)$ to the processed data matrix $\phi_P(\bX;\btheta_P)$. Our choice of kernel function is a mixture of Gaussians,
\begin{equation}
    k(\bz_i,\bz_j) = \frac{1}{T}\sum_{t=1}^{T} \exp\lp-\frac{\lnorm \bz_i-\bz_j \rnorm_2^2}{2 \sigma_t^2}\rp\text{,}
    \label{eq:gauss_mixture}
\end{equation}
where $\{\sigma_t\}_{t=1}^{T}=\{1,2,4,8,16\}$. This is the same setting as in \cite{li2017mmd,li2015generative}, where MMD was utilized for training GANs. Similar to those works, our main reason for choosing a mixture is the reduced burden of parameter tuning, in addition to the stronger representation capacity of a mixture of Gaussians.

\textbf{KDI}: We utilize the privacy objective in \eqref{eq:KDI}, which leads to the privacy loss function
\begin{multline}
    L_P(\bX,\bP;\btheta_P) = \trace \Big( \lp \Kbar^2(\bX;\btheta_D)+\rho  \Kbar(\bX;\btheta_D) \rp^{+} \\ \Kbar(\bX;\btheta_D) \bP \bP^\top \Kbar(\bX;\btheta_D) \Big) \text{,}
    \label{eq:KDI_obj}
\end{multline}
where $\Kbar(\bX;\btheta_D)=\bC\bK(\bX;\btheta_D)\bC$, with $\bC=\bI-\frac{1}{N}\bone$ being the centering matrix. $\bK(\bX;\btheta_D)$ is the same kernel matrix used in the definition of the MMD objective. We set the ridge regularizer $\rho$ to $10^{-4}$ in our experiments.

\textbf{WDN}: We use the inverse of the Wasserstein Discriminator Network objective from \eqref{eq:wd_multi_linear}. Let the privacy label matrix $\bP$ contain the one-hot encodings of privacy classes in its rows. Let $\bp_l$ denote the $l^{th}$ column of the $N\times L$ matrix $\bP$ and $\overline{\bp}_l=\rone-\bp_l$, the privacy loss function is given by
\begin{multline}
    L_P(\bX,\bP,\btheta_D;\btheta_P) \\= -\sum_{l=1}^{L} \frac{N_l}{N} \Bigg( \frac{1}{N_l} {\phi_D}(\phi_P(\bX;\btheta_P);\btheta_D) \bp_l \\- \frac{1}{\overline{N}_l}\phi_D(\phi_P(\bX;\btheta_P);\btheta_D) \overline{\bp}_l \Bigg) \text{,}
    \label{eq:WDN_obj}
\end{multline}
where $N_l=\rone^\top \bp_l$,  $\overline{N}_l=\rone^\top \overline{\bp}_l$.

\textbf{LSDN}: We use the adversarial training objective from \eqref{eq:lsdn_priv_loss}, which leads to the privacy loss function
\begin{multline}
    L_P (\bX,\bP,\btheta_D;\btheta_P) \\= \frac{1}{N}\lnorm \phi_D(\phi_P(\bX;\btheta_P);\btheta_D)-\frac{1}{N} \bP^\top \rone \rone^\top \rnorm_F^2 \text{.}
    \label{eq:LSDN_obj}
\end{multline}

Note that the MMD \eqref{eq:MMD_obj} and KDI \eqref{eq:KDI_obj} objectives are free from the parameters $\btheta_D$. This is because these closed-form statistics do not require an explicit discriminator network to be defined. For the WDN \eqref{eq:WDN_obj} and LSDN \eqref{eq:LSDN_obj} objectives to work, we have to optimize a privacy discriminator jointly with the private and public sphere networks. 

Another important distinction is the settings in which these four privacy losses are definable. Namely, KDI \eqref{eq:KDI_obj} and LSDN \eqref{eq:LSDN_obj} losses are defined for any arbitrary label matrix $\bP$, whereas MMD \eqref{eq:MMD_obj} and WDN \eqref{eq:WDN_obj} losses are only defined if $\bP$ contains one-hot encodings of class labels. This is because MMD and WDN are only defined for discrete variables and are not suitable for continuous variables.

\subsubsection{The Objectives for The Privacy Discriminator} 
Below are the discriminator objectives we utilize for optimizing $\btheta_D$.

\textbf{WDN}: The overall loss of this network is the sum of the linear discriminator loss $L_D$ \eqref{eq:wd_multi_linear} and the Lipschitz penalty $L_R$ \eqref{eq:wd_multi_lipschitz}. If the privacy label matrix $\bP$ contains the one-hot encodings of privacy classes in its rows, these losses can be written as
\begin{align}
     &L_D(\bX,\bP,\btheta_P;\btheta_D) \nonumber \\&= \sum_{l=1}^{L} \frac{N_l}{N} \Big( \frac{1}{N_l} {\phi_D}(\phi_P(\bX;\btheta_P);\btheta_D) \bp_l \nonumber\\& \qquad \qquad \qquad - \frac{1}{\overline{N}_l}\phi_D(\phi_P(\bX;\btheta_P);\btheta_D) \overline{\bp}_l \Big) \text{,} \\
    &L_R(\bX,\bP,\btheta_P;\btheta_D) \nonumber \\ &=\frac{1}{N}\sum_{i=1}^{N} \lp \lnorm \nabla_{\bphi_p(\bx_i;\btheta_P)} \lp \bs_i^\top{\phi_D}(\bphi_p(\bx_i;\btheta_P);\btheta_D)\rp \rnorm_2 - 1  \rp^2 \text{,}
\end{align}
where $\bp_l$ refers to the $l^{th}$ column of $\bP$, while $\bs_i$ refers to the $i^{th}$ row of $\bP$, $\overline{\bp}_l=\rone-\bp_l$, $N_l=\rone^\top \bp_l$,  $\overline{N}_l=\rone^\top \overline{\bp}_l$. The overall WDN objective is then given by
\begin{align}
    &L_{Disc} (\bX,\bP, \btheta_P;\btheta_D) = L_D(\bphi_p(\bX;\btheta_P),\bP,\btheta_P;\btheta_D) \nonumber\\ &\qquad \qquad \qquad \qquad+\lambda_{R} L_R(\bphi_p(\bX;\btheta_P),\bP,\btheta_P;\btheta_D) \text{.}
    \label{eq:wdn_disc_obj}
\end{align}

We use $\lambda_R=10$ in our experiments, which is consistent with the work that proposed this regularizer \cite{gulrajani2017improved}.

\textbf{LSDN}: This network minimizes the squared error in \eqref{eq:lsdn_loss} with no regularizer, so the LSDN objective is given by 
\begin{equation}
    L_{Disc} (\bX,\bP, \btheta_P;\btheta_D) = \frac{1}{N}\lnorm \phi_D(\phi_P(\bX;\btheta_P);\btheta_D)-\bP^\top \rnorm_F^2 \text{.}
    \label{eq:lsdn_disc_obj}
\end{equation}

\subsection{The Training Procedure} \label{sec:end_to_end_procedure}

\begin{algorithm}[t]
\caption{Utility and Privacy Maximizing Model Training}
\begin{algorithmic}
\small
\State \textbf{Inputs:} Training data: $(\bX,\bY,\bP)$; privacy parameter $\lambda_P$, step size $\alpha$, $batch\_size$; private and public sphere network architectures: $\phi_P$, $\phi_U$, [a privacy discriminator architecture $\phi_D$, step size $\alpha_D$].
\State \textbf{Output:} Optimized private sphere network $\phi_P(\cdot,\btheta_P)$.
\State -- Select a privacy objective, if MMD or KDI is chosen, ignore the parts in square brackets.
\If {MMD} $L_P$ is in \eqref{eq:MMD_obj}.
\ElsIf {KDI} $L_P$ is in \eqref{eq:KDI_obj}.
\ElsIf {WDN} $L_P$ is in \eqref{eq:WDN_obj}, $L_{Disc}$ is in \eqref{eq:wdn_disc_obj}.
\ElsIf {LSDN} $L_P$ is in \eqref{eq:LSDN_obj}, $L_{Disc}$ is in \eqref{eq:lsdn_disc_obj}.
\EndIf
\State -- Initialize the private and public sphere network parameters $\btheta_P$, $\btheta_U$, [initialize privacy discriminator parameters $\btheta_D$].
\If {$\phi_P$ is a CNN ending with a dense layer with parameters $\bW\subset\btheta_P$} $L_P \leftarrow L_P + L_O$ where $L_O$ is in \eqref{eq:orthonormality}. \EndIf
 \Repeat
  \For {$b=1, \ldots, \lfloor N/batch\_size \rfloor$} 
   \State Extract a mini-batch $(\bX',\bY',\bP') \subset (\bX,\bY,\bP)$\;
   \State $\btheta_P \leftarrow \btheta_P - \alpha\nabla_{\btheta_P}\left(L_U +\lambda_P L_P\right)(\bX',\bY',\bP',\btheta_U,[\btheta_D])$
   \State $\btheta_U \leftarrow \btheta_U - \alpha\nabla_{\btheta_U} L_U(\bX',\bY',\btheta_P)$ where $L_U$ is in \eqref{eq:util_loss}
   %\State  $\nabla \boldsymbol{\theta} = \frac{\partial \text{NysDI}(\bX',\bY';\boldsymbol{\theta})}{\partial \boldsymbol{\theta}}$
   \State [$\btheta_D \leftarrow \btheta_D-\alpha_D\nabla_{\btheta_D}L_{Disc}(\bX',\bP',\btheta_P)$]
  \EndFor
 \Until MMD/KDI: $L_U -\lambda_P L_P$ converges OR WDN/LSDN: A preset number of epochs have passed
\end{algorithmic}
\label{alg:Algorithm}
\vspace{-3pt}
\end{algorithm}

A summary of our privacy enhancing training methodology is provided in Algorithm \ref{alg:Algorithm}. We begin describing it by writing the overall network losses minimized by the private sphere network, public sphere network and the privacy discriminator, respectively. The private sphere network minimizes a linear combination of the utility loss $L_U$ \eqref{eq:util_loss} and one of the MMD \eqref{eq:MMD_obj}, KDI \eqref{eq:KDI_obj}, WDN \eqref{eq:WDN_obj}, LSDN \eqref{eq:LSDN_obj} privacy losses $L_P$,
\begin{multline}
    L_{Pri}(\bX,\bY,\bP,\btheta_U,\btheta_D;\btheta_P) \\= L_U(\bX,\bY;\btheta_P,\btheta_U) + \lambda_P L_P(\bX,\bP,\btheta_D;\btheta_P) \text{,}
    \label{eq:priv_sphere_loss}
\end{multline}
where $\lambda_P$ controls the importance of the privacy objective. We shall vary this parameter in our experiments to showcase the utility/privacy trade-offs with different objectives and network architectures. If we use CNNs as our private and public sphere networks, we also add dense subspace projection layers to the intersection of the private and public spheres as in \eqref{eq:cnn_funnel}. In this case, we also add the orthonormality penalty \eqref{eq:orthonormality} to \eqref{eq:priv_sphere_loss}.

The public sphere network only minimizes the utility loss $L_U$ \eqref{eq:util_loss},
\begin{equation}
    L_{Pub}(\bX,\bY,\btheta_P;\btheta_U) = L_U(\bX,\bY;\btheta_P,\btheta_U) \text{.}
\end{equation}

Finally, if we utilize the WDN \eqref{eq:WDN_obj} or the LSDN  \eqref{eq:LSDN_obj} loss as our privacy objective, we train a privacy discriminator network that minimizes $L_{Disc}$ as given by \eqref{eq:wdn_disc_obj} or \eqref{eq:lsdn_disc_obj}, respectively.

We use stochastic gradient methods to jointly optimize these networks, therefore, the loss gradients are computed based on mini-batches, which are subsets of the training set $(\bX,\bY,\bP)$. Due to the adversarial nature of the private sphere network with respect to the privacy discriminator, we keep the step size of the private sphere network smaller than the step size of the privacy discriminator. This is to ensure that the privacy discriminator can adapt to changes in the private sphere network. Of course, this is no concern when we are using the MMD and KDI objectives. 

We found that applying regularizers like drop-out and batch-normalization to private and public sphere networks can help generalize to the utility prediction task. However, these methods should not be applied to the privacy discriminator, since this alters the privacy objectives, which can significantly lower the privacy performance of the private sphere network. Therefore, we apply drop-out to the private and public sphere networks during training, except for the narrow, funneling layer at the intersection of these networks.

\section{Experiments}

We perform two sets of experiments to verify the effectiveness of the model architectures and learning objectives presented in Section \ref{sec:methodology}. The first set of experiments is concerned with optimizing linear projections in the private sphere, and the second is concerned with optimizing CNN mappings. To facilitate meaningful comparisons among the four privacy objectives, we consider settings with discrete privacy variables. The utility/privacy trade-offs are obtained by gradually increasing the privacy parameter $\lambda_P$ in \eqref{eq:priv_sphere_loss}. We generally found that exploring the parameter ranges $[2^{-10}, 2^{10}]$ for MMD and WDN, $[2^{-10}, 1]$ for KDI and $[2^{-4}, 2^{12}]$ for LSDN privacy objectives to be sufficient to capture the trade-offs between minimal and maximal privacy settings.

We use the Adam optimizer \cite{kingma2014adam} with a batch size of 500 throughout our experiments. We set the step size to $10^{-3}$ while optimizing the private and public sphere networks with the MMD \eqref{eq:MMD_obj} and KDI \eqref{eq:KDI_obj} privacy objectives. We use a step size of $10^{-3}$ for the privacy discriminator and a step size of $10^{-4}$ for the public and private sphere networks while using the WDN \eqref{eq:WDN_obj} and LSDN \eqref{eq:LSDN_obj} privacy objectives. The learning rates are periodically reduced by a factor of $10^{-1}$ during training, which we generally found to improve the utility/privacy performances. When utilizing MMD or KDI, lowering of the learning rates is done when the overall objective fails to decline. When utilizing WDN or LSDN, a fixed schedule of 250 epochs is used.

\subsection{Learning Privacy Enhancing Linear Projections}\label{subsec:Learning Privacy Enhancing}
We start our experiments with one of the simplest data processing methods on the user side, which is linear projections. We use the HAR \cite{anguita2013public} and MHEALTH \cite{banos2014mhealthdroid} datasets in these experiments. We perform 10 randomized experiments with different splits of these data and report the average performances.

The HAR data contains $561$-feature samples from $30$ users performing $6$ activities. We split the $10299$ samples into training and test sets with $80:20$ ratios, and make sure the training and test sets contain the same proportion of samples from each user. The original split of this dataset ensures that training and test sets have non-overlapping users, hence, it is possible to predict user activities without using information related to user identity. For this reason, we consider the \emph{activity recognition as the utility prediction task}, and \emph{identity recognition as the privacy prediction task}. We set the linear projection dimensions to $50$. We found this dimensionality to be sufficient for maintaining the utility performance on this dataset.

The MHEALTH data contains $23$-feature samples from $10$ users performing $12$ activities. We extract $12000$ frames ($100$ frames per activity per user) for our training set and $3000$ frames ($25$ frames per activity per user) for our test set. We ensure that there is at least a $400$ frame separation between the training and test sets. Once again, we consider the \emph{activity recognition as the utility prediction task}, and \emph{identity recognition as the privacy prediction task}. We set the linear projection dimensions to $10$ for this data. While this number of dimensions was found to be restrictive in the sense that it reduces the utility performance, we chose it due to the fact that the original number of features is small on this data. 

Linear projections for preserving privacy were optimized using Discriminant Analysis related methods in previous works. To showcase the improvements we can get over these, we build a system to optimize privacy enhancing/utility preserving linear projections in progressive stages. These are listed below.

\begin{itemize}
    \item \textbf{Discriminant Utility-Cost Analysis (DUCA)} \cite{kung2017compressive,chanyaswad2016discriminant}: Optimizes linear projections based on the objective
    \begin{align}
     &\underset{\bW \colon \bW^\top \lp \Xbar \Xbar^\top + \rho \bI \rp \bW = \bI}{\text{maximize}} \trace \Bigg( \bW^\top \Big(\Xbar \bY \bY^\top \Xbar^\top -\rho' \bI \nonumber\\& \qquad \qquad \qquad \qquad \qquad \ - \lambda_P \Xbar \bP \bP^\top \Xbar^\top \Big)  \bW \Bigg) \text{,}
     \label{eq:exp_duca}
    \end{align}
    where $\Xbar=\bX \bC$ is the centered data matrix. The ridge regularizers $\rho,\rho'$ are set according to \cite{kung2017compressive,chanyaswad2016discriminant} and the optimal solutions are found via generalized eigenvalue decomposition. The significance of this method is that it uses the linear variant of the DI criterion for both the utility objective and the privacy objective.
    \item \textbf{DUCA-MMD}: We use the utility part of the DUCA objective \eqref{eq:exp_duca}, but replace the privacy objective with MMD, that is, by considering $\phi_P(\bX;\btheta_P)=\bW^\top \bX$, this system minimizes the loss function
    \begin{multline}
     -\trace\lp \lp \bW^\top\Xbar \Xbar^\top\bW + \rho \bI \rp^{-1} \bW^\top \Xbar \bY \bY^\top \Xbar^\top \bW \rp \\+ \lambda_P L_P(\bX,\bP;\btheta_P) \text{,}
    \end{multline}
    where $\btheta_P \coloneqq \bW$ and $L_P$ is the MMD objective in \eqref{eq:MMD_obj}. We use the equivalent RR predictor as the public sphere network to minimize the utility loss, since we established the equivalence between DI and minimum RR loss in Section \ref{sec:kdi}.
    \item \textbf{NN-MMD}: We use a public sphere network minimizing the cross-entropy (CE) loss in addition to the private sphere network $\phi_P(\bX;\btheta_P)=ReLU(\bW^\top \bX+\bb)$ minimizing the MMD version of \eqref{eq:priv_sphere_loss}. Note that the addition of the bias term and rectified linear units is computationally negligible, but they can add privacy benefits due to ReLU removing some projection directions from the private sphere output. The public sphere network $\phi_U(\cdot;\btheta_U)$ consists of one hidden layer with 500 units and an output layer. The optimization of these networks was performed as we described in Section \ref{sec:end_to_end_procedure}.
    \item \textbf{NN-KDI/WDN/LSDN}: We use the same private and public sphere networks as \textbf{NN-MMD}, but replace the privacy loss with one of the three alternatives described in Section \ref{subsubsec:Priv Obj}. The privacy discriminators $\phi_D(\cdot;\btheta_D)$ of the WDN and LSDN models consist of one hidden layer with 1024 units and an output layer.\footnote{In addition to the methods described here, we considered the approach in \cite{feng2019learning}, where a linear discriminator network is used to create the WDN objective. As we found that linear discriminators lead to poor privacy results compared to their non-linear counterparts, however, we only report the results from the WDN and LSDN objectives obtained with non-linear discriminators.}
    \item \textbf{NN-(RF)MMD}: The methodology is the same as \textbf{NN-MMD}, but instead of the standard mixture of Gaussian kernels described in \eqref{eq:gauss_mixture}, a Random Fourier approximation of the Gaussian kernel is utilized to define the MMD objective as in \cite{louizos2015variational}. Though, we choose the approximation dimensionality to be $1000$ (instead of 500) to give the resulting objective more representation capacity to compete with our standard MMD objective.
\end{itemize}

\begin{figure*}[t]

\begin{minipage}[b]{0.49\linewidth}
  \centering
  \centerline{\includegraphics[width=9cm]{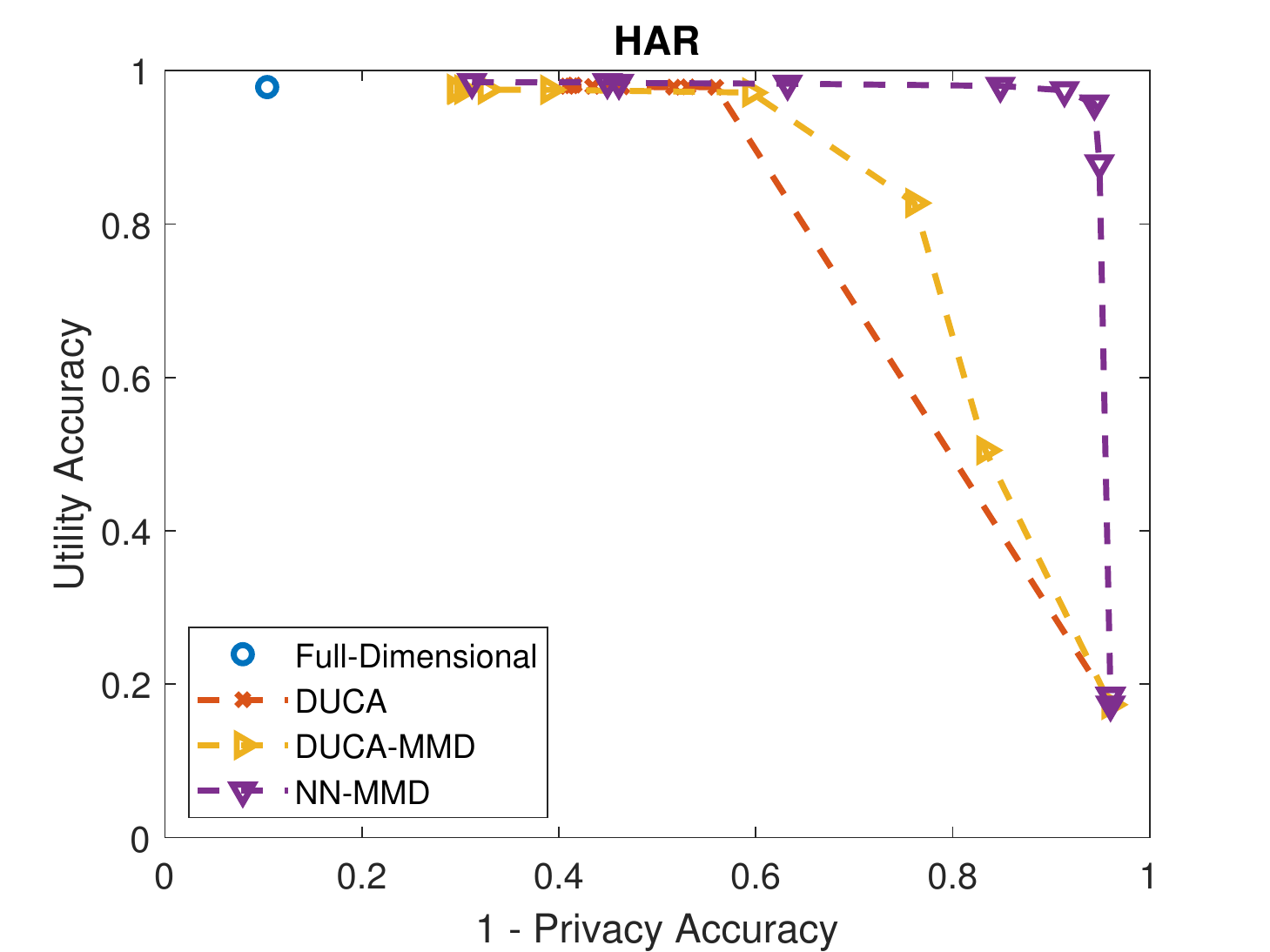}}
\end{minipage}
\hfill
\begin{minipage}[b]{0.49\linewidth}
  \centering
  \centerline{\includegraphics[width=9cm]{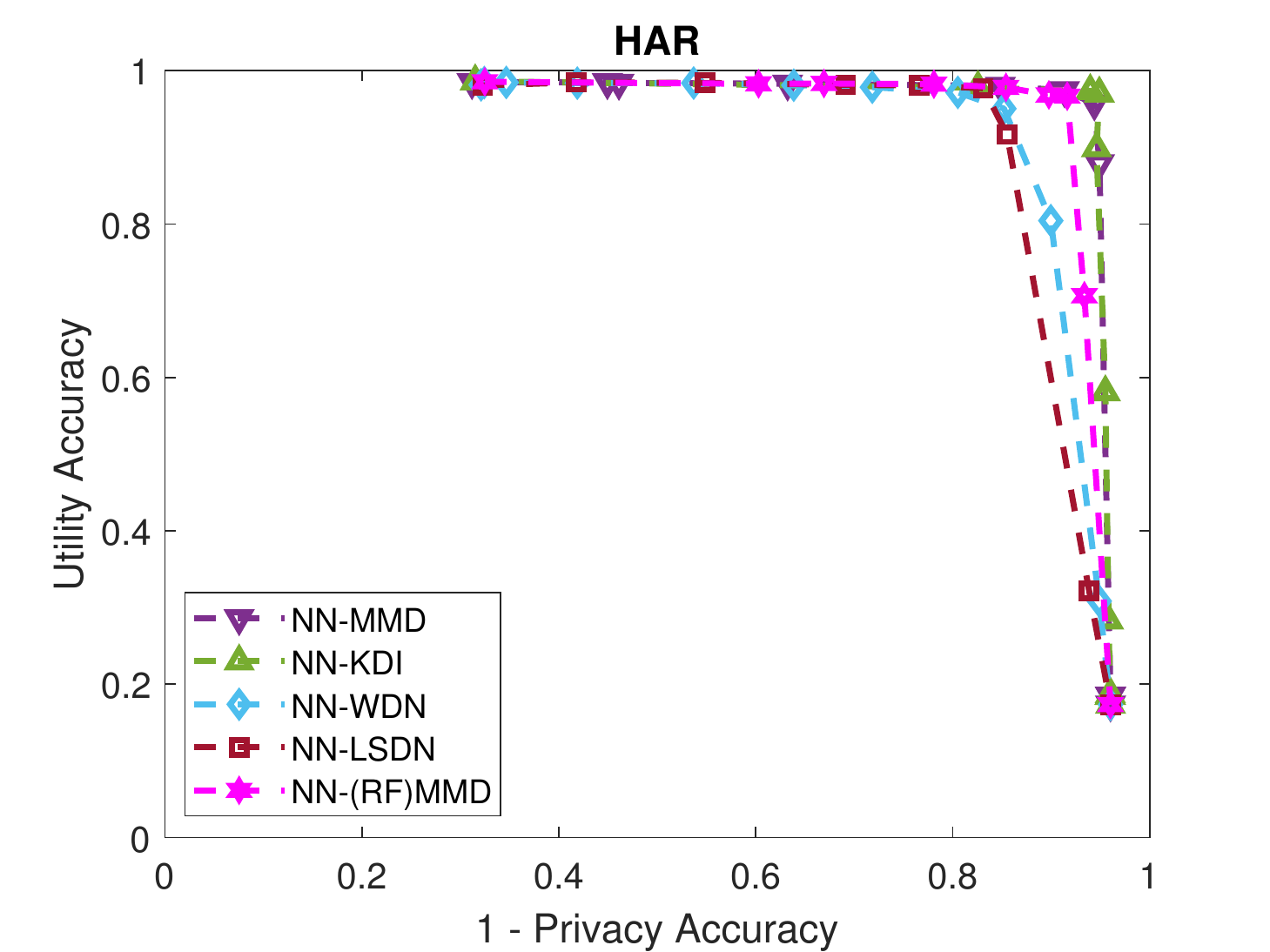}}
\end{minipage}
\vfill
\vspace{2mm}
\begin{minipage}[b]{0.49\linewidth}
  \centering
  \centerline{\includegraphics[width=9cm]{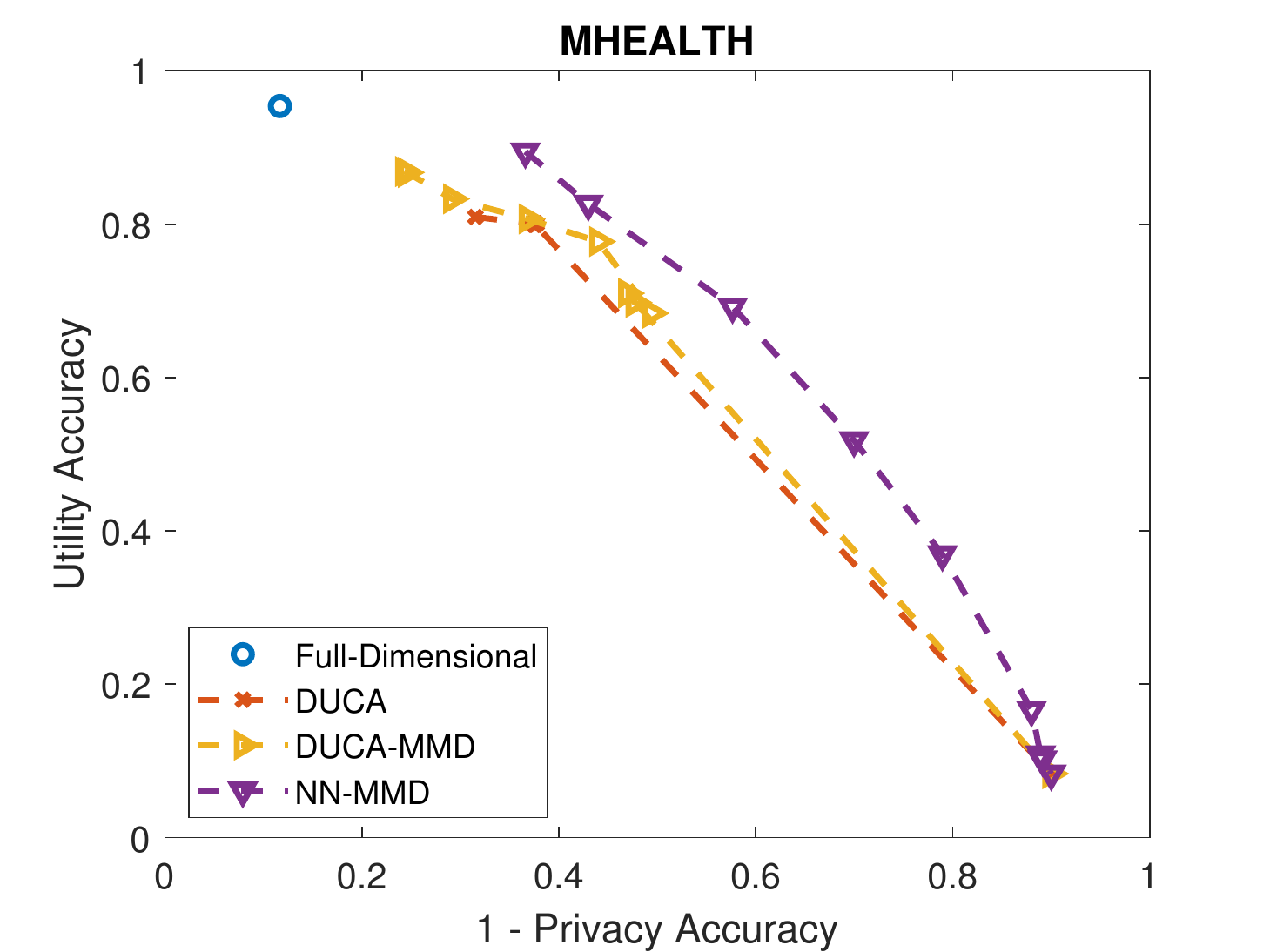}}
\end{minipage}
\hfill
\begin{minipage}[b]{0.49\linewidth}
  \centering
  \centerline{\includegraphics[width=9cm]{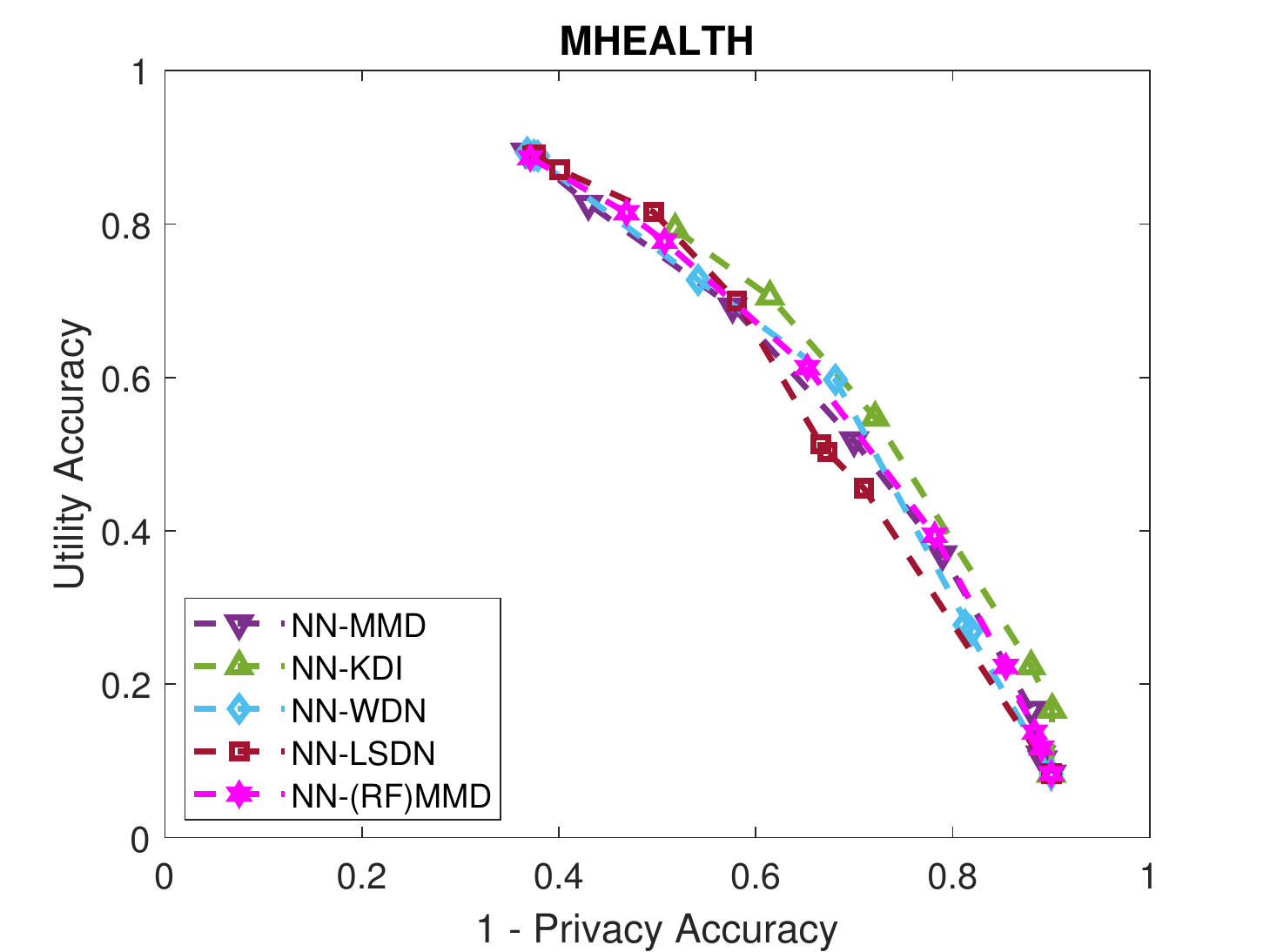}}
\end{minipage}
\caption[Utility-Privacy Curves of Optimized Linear Projections and DNNs]{Utility vs. Privacy trade-off curves obtained from the presented linear projections applied to the HAR (top) and MHEALTH (bottom) datasets. Left figures compare our system with alternative linear methods, and the right figures compare our systems with each other using different privacy objectives.}
\label{fig:har_end1}
\end{figure*}

To test the privacy performances of all the models, we utilize the following predictors to represent the adversary, who might want to infer sensitive information from the data:
\begin{itemize} 
\item Linear SVM, whose $l_2$ penalty is optimized separately for every experiment via 5-fold cross-validation.
\item RBF Kernel SVM, whose $l_2$ penalty and kernel bandwidth are optimized separately for every experiment via 5-fold cross-validation.
\item Random Forest (RF) predictor with 250 trees.
\item Neural Network (NN) predictor minimizing softmax cross-entropy, whose architecture mimics the privacy discriminator network.
\end{itemize} 
These adversarial models are trained on the data after the data desensitizing feature mapping $\phi_P(\bX;\btheta_P)$ is applied. In each experiment, we consider the adversary's performance to be \emph{the maximum accuracy achieved} among these predictors. 

We use RF predictors to test the utility performances of \textbf{DUCA} and \textbf{DUCA-MMD} models, as they do not optimize utility predictors themselves. The utility performances of the \textbf{NN-MMD/KDI/WDN/LSDN} models were measured as the performances of their public sphere predictors, which minimize the softmax cross-entropy loss for the utility prediction task.

Figure \ref{fig:har_end1} shows our comparisons among the methods considered. The full-dimensional points represent the utility and privacy performances when the original data are released. The methods we propose improve the privacy performances over this scenario, but they eventually reduce both the utility and privacy performances down to random guessing (at high privacy parameter settings).

The comparison between \textbf{DUCA} and \textbf{DUCA-MMD} demonstrates the potential gain from utilizing non-linear privacy objectives even while optimizing linear projections. \emph{We see that \textbf{DUCA} does a relatively poor job at desensitizing the data, mainly because it only captures the linear correlations between variables}. Due to this limitation, \textbf{DUCA} becomes insensitive to changes in the privacy parameter after all the linear correlations between the projected data and private variables are removed.\footnote{We also consider random guessing the utility and privacy variables as part of the \textbf{DUCA} trade-off curves, because it represents choosing not to share any information at all.} \emph{\textbf{DUCA-MMD}, on the other hand, utilizes the MMD privacy objective, which is sensitive to non-linear correlations between the projected data and private variables}. Thus, it is able to remove more private information at high privacy settings. This results in a significantly better utility/privacy trade-off on HAR data, where we learn a $561 \times 50$ projection matrix (a relatively high degree of freedom), though the improvement is very limited on MHEALTH, where we learn a $23 \times 10$ projection matrix (a relatively low degree of freedom).

Finally, our \textbf{NN-MMD} method achieves an extremely desirable utility/privacy trade-off on HAR data by being able to lower the privacy performance close to random guessing without sacrificing much utility performance. Although linear projections prove too restrictive on MHEALTH data to achieve ideal utility/privacy performances, \textbf{NN-MMD} also improves the utility/privacy trade-off significantly here.\footnote{Early experiments revealed that the utility/privacy trade-offs can be improved by adding more dense layers into the private sphere network. However, this might place the bulk of the computational burden on the user side, which goes against the design philosophy behind this paper. Hence, we elected to restrict the private sphere networks to contain a single, narrow dense layer in our experiments.} This is thanks to the joint optimization of the public and private sphere networks in addition to the more general utility and privacy objectives being utilized. Among our own methods, \textbf{NN-MMD} and \textbf{NN-KDI} achieve statistically similar performances, with \textbf{NN-WDN} being a close second and \textbf{NN-LSDN} performing worse than the other three methods on HAR data. On the other hand, these four methods achieve very similar performances on MHEALTH with \textbf{NN-KDI} having a slight edge. 

When we replace the mixture of Gaussian kernels with a Random Fourier approximation, we seem to achieve slightly worse privacy performance on HAR as demonstrated by the comparison between \textbf{NN-MMD} and \textbf{NN-(RF)MMD}. This might be explained by the stronger representation capacity of a mixture of Gaussian kernels compared to a Random Fourier approximation, even though the two produce extremely similar results on MHEALTH.

The experiments in this section involved only a single, narrow processing layer in the private sphere. We show in Section \ref{subsec:Learning CNNs} that, by adding more processing layers into the private sphere, the utility/privacy performances can be improved further.

\subsection{Learning Privacy Enhancing CNNs}\label{subsec:Learning CNNs}

\begin{figure*}[t]
\begin{minipage}[b]{0.49\linewidth}
  \centering
  \centerline{\includegraphics[width=9cm]{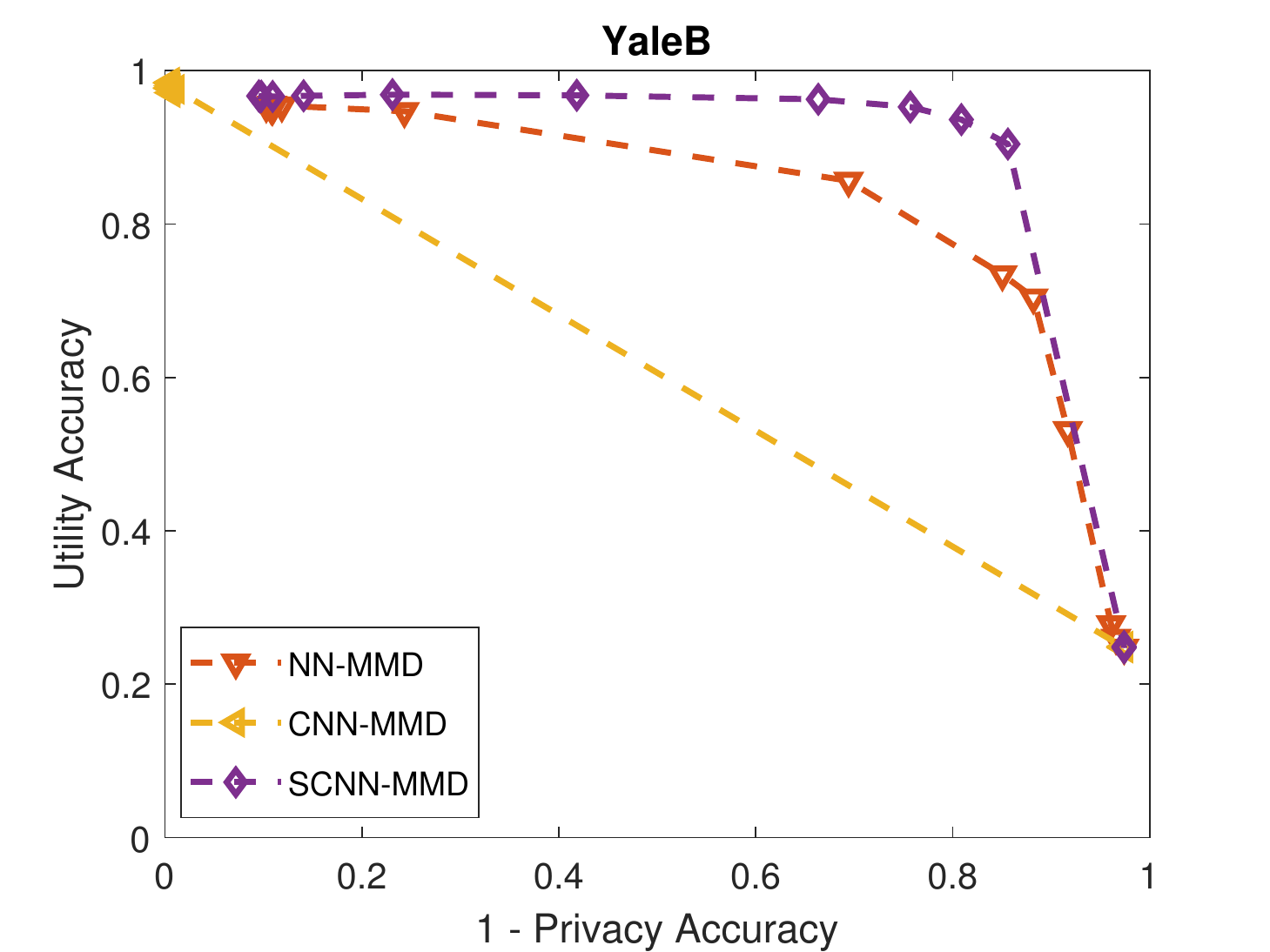}}
\end{minipage}
\hfill
\begin{minipage}[b]{0.49\linewidth}
  \centering
  \centerline{\includegraphics[width=9cm]{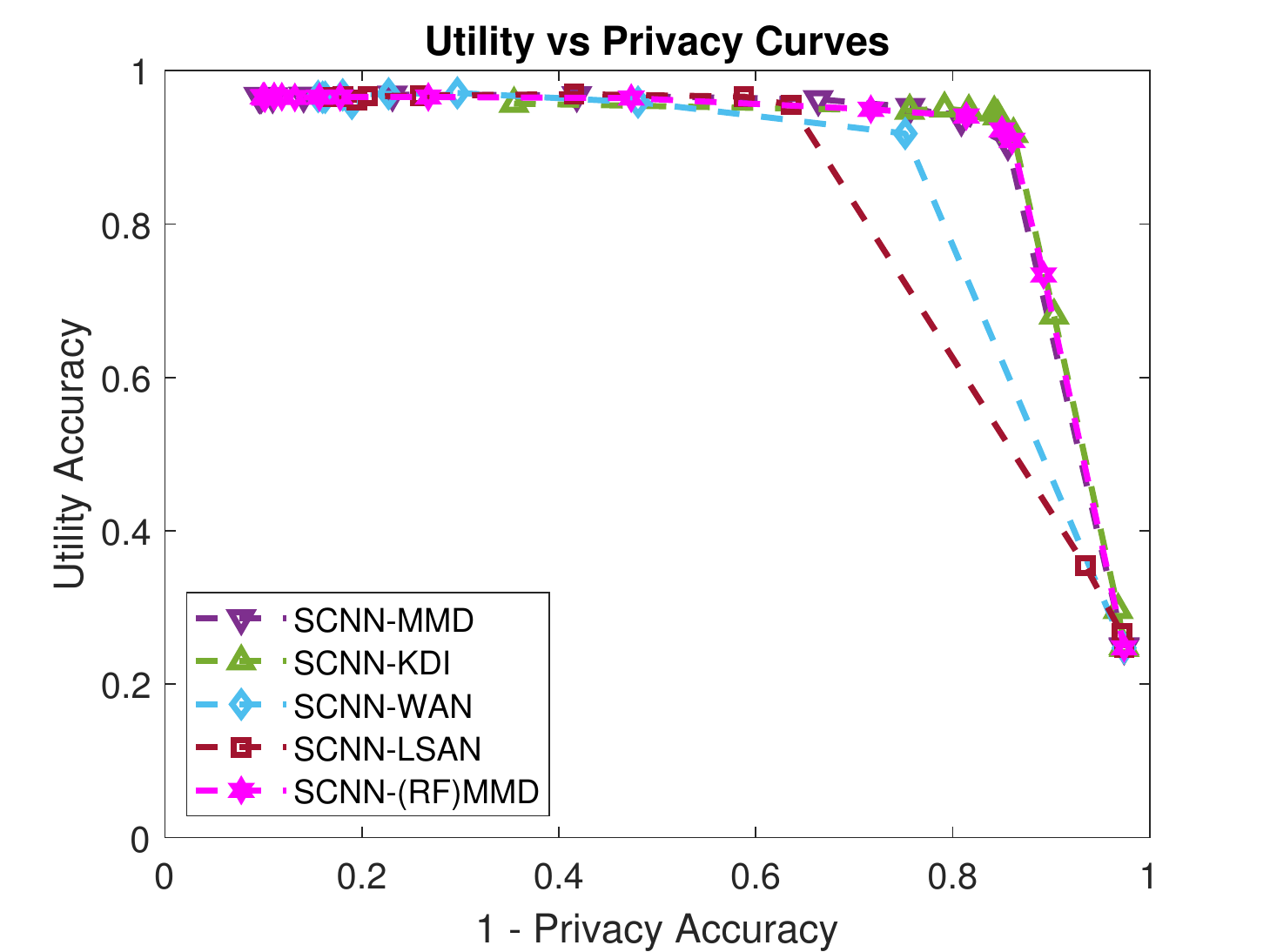}}
\end{minipage}
\vfill
\vspace{2mm}
\begin{minipage}[b]{0.49\linewidth}
  \centering
  \centerline{\includegraphics[width=9cm]{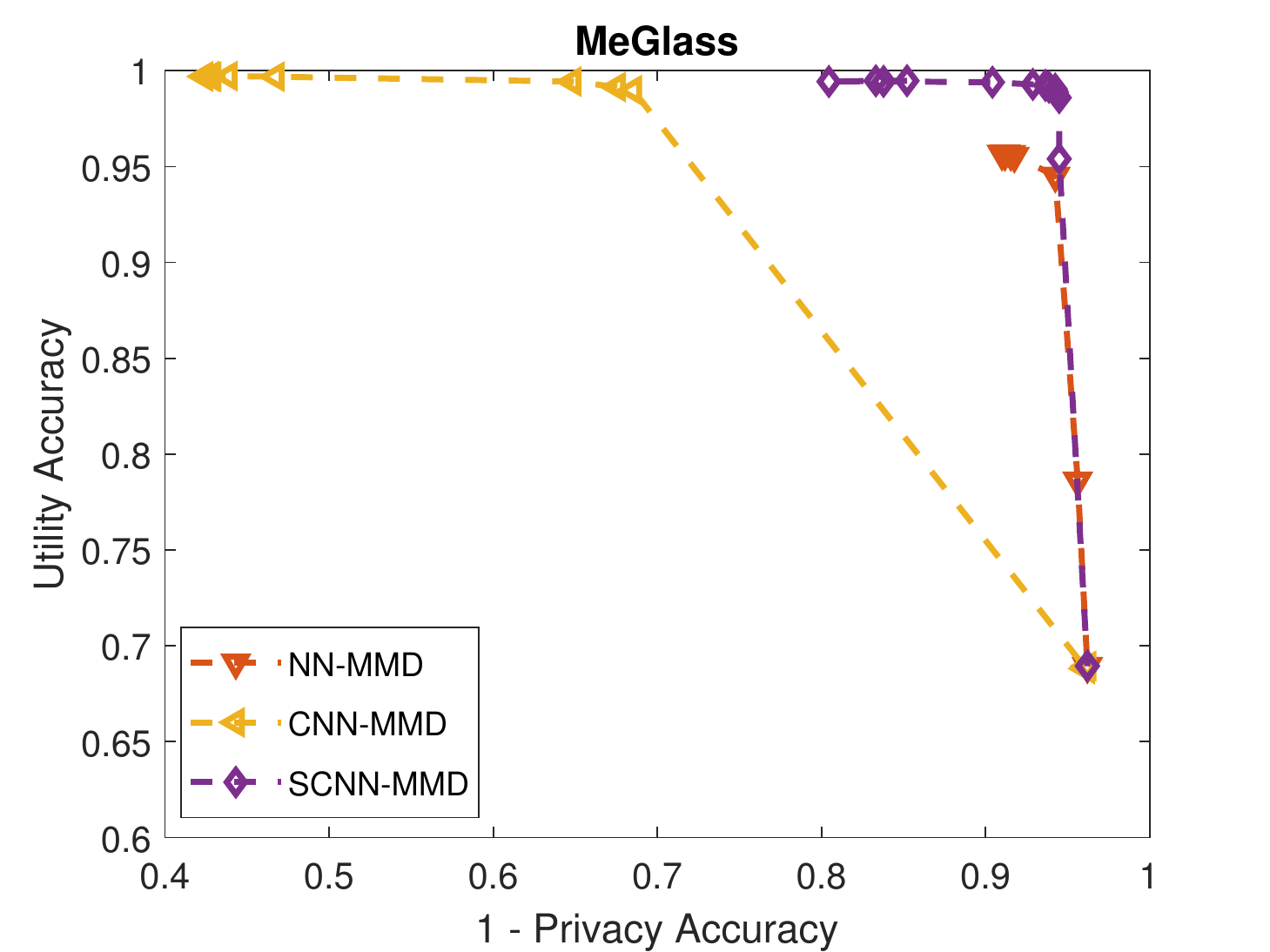}}
\end{minipage}
\hfill
\begin{minipage}[b]{0.49\linewidth}
  \centering
  \centerline{\includegraphics[width=9cm]{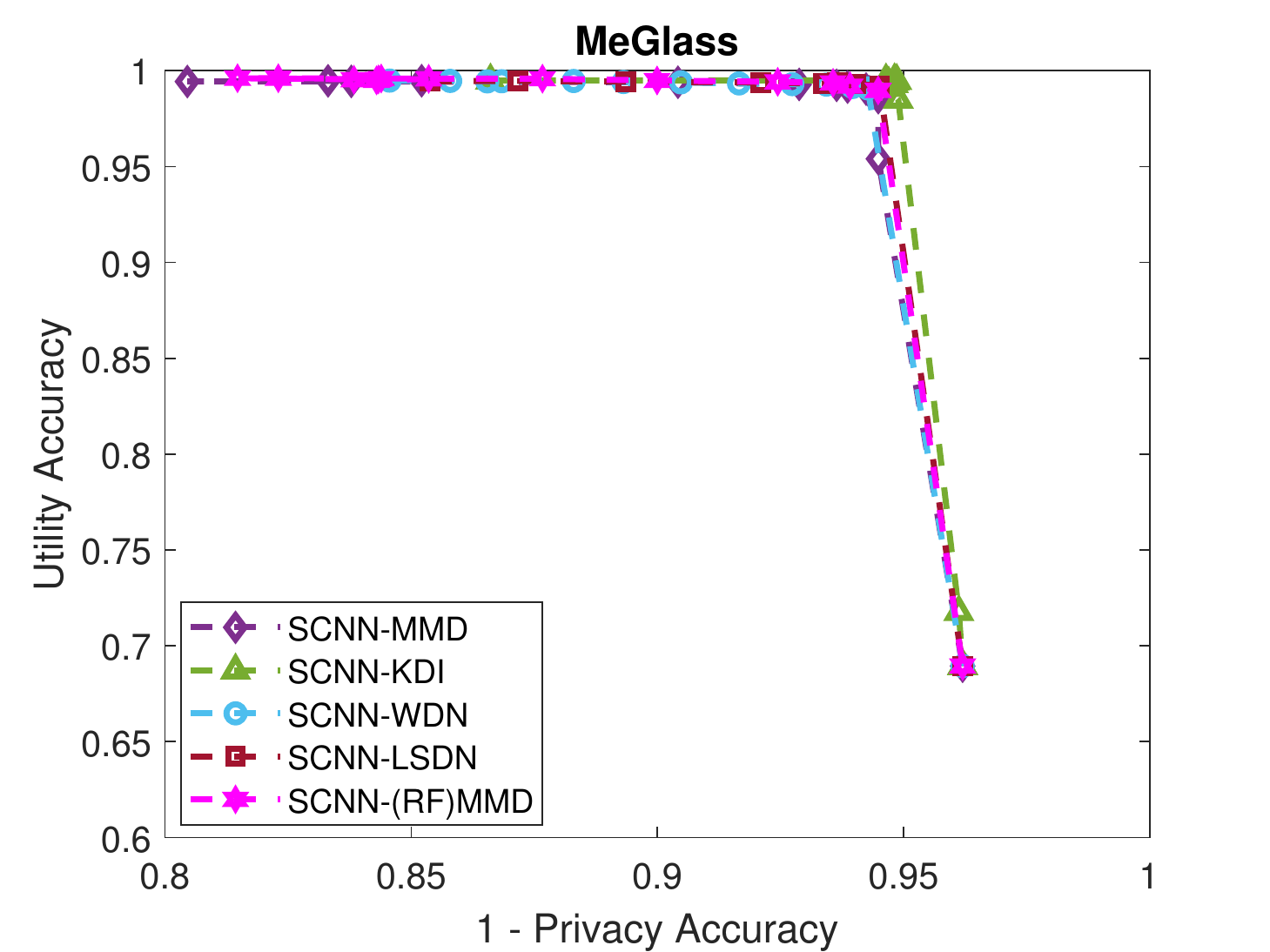}}
\end{minipage}
\caption[Utility-Privacy Curves of Optimized DNNs and CNNs]{Utility vs. Privacy trade-off curves obtained from the presented structures applied to the YaleB and MeGlass datasets. Left figures compare different architectures using the same objectives, and the right figures compare the same architectures using different privacy objectives.}
\label{fig:yaleb_end1}
\end{figure*}

In this section, we consider adding CNN layers both before and after the privacy enhancing subspace projections. We use the extended YaleB \cite{lee2005acquiring} and MeGlass \cite{guo2018face} datasets for these experiments, both of which consist of face images. Once again, we perform 10 randomized experiments with different splits of these data and report the average performances.

YaleB data contains $2414$ face images from $38$ individuals. We split this data into training and test sets with a $80:20$ ratio, making sure the sets contain the same proportion of samples from each individual. The face images are reshaped to be $32\times32$, and we randomly divide the individuals into $4$ groups to create a \emph{utility prediction task that corresponds to accessing coarser granular information about user identity}. Hence, we treat the \emph{prediction of the user group as the utility prediction task} and the \emph{prediction of the individual identity as the privacy prediction task}. Notice that \emph{hiding all the sensitive information in this setting destroys all the utility information as well}, so privacy cannot be improved beyond a certain level without sacrificing utility performance.

MeGlass data contains $47917$ face images from $1565$ individuals, each of whom have at least two photos with glasses and two photos without glasses. We randomly select $500$ of these individuals for each experiment, which leads to roughly $15600$ samples per experiment. We split these samples into training and test sets with a $80:20$ ratio, keeping the proportion of samples from each individual consistent. To reduce the number of classes in the adversarial learning task, we randomly split the $500$ individuals into $100$ groups of $5$ (the number of samples can differ across groups, but no group becomes larger than $4.5 \%$ of the data). Accordingly, we treat \emph{glass detection as the utility prediction task} and \emph{detection of an individual's group as the privacy prediction task}. Based on the original purpose of this dataset, \emph{we know that glass detection can be performed independently from a person's identity, hence, we expect our methods to achieve near ideal utility/privacy trade-offs on this data}.

We build upon the system we designed in Section \ref{subsec:Learning Privacy Enhancing}, and develop a privacy enhancing CNN architecture in progressive stages. The projection dimensions are set to 50 across the models (i.e., the narrow, funneling layer has 50 units in all the models where it exists). While the CNNs considered in this section are very rudimentary compared to the state of the art, we found that they are sufficient to perform well on these simpler learning tasks on relatively small benchmark datasets. The models we consider are listed below.

\begin{itemize}
    \item \textbf{NN-MMD}: This has the same model structure and objectives as the \textbf{NN-MMD} in Section \ref{subsec:Learning Privacy Enhancing}. We have a private sphere network $\phi_P(\bX;\btheta_P)=ReLU(\bW^\top \bX+\bb)$ and a public sphere network $\phi_U(\cdot;\btheta_U)$, which consists of one hidden layer with 1024 units and an output layer. 
    \item \textbf{CNN-MMD}: This optimizes the same model objectives as \textbf{NN-MMD}, but both the private and public sphere models are replaced by CNNs. The private sphere network $\phi_P(\bX;\btheta_P)$ consists of one convolutional layer with $32$ $3\times3$ filters followed by $2\times 2$ max-pooling. The public sphere network $\phi_U(\cdot;\btheta_U)$ consists of one convolutional layer with $64$ $3\times3$ filters followed by $2\times 2$ max-pooling, one dense layer with $1024$ units and an output layer. No subspace projection takes place between private and public spheres.
    \item \textbf{SCNN-MMD}: We add a subspace projection layer \eqref{eq:cnn_funnel} at the intersection of the private and public sphere networks. Accordingly, the new private sphere network becomes $\phi_P(\bX;\btheta_P)=ReLU(\bW^\top \phi_P'(\bX;\btheta_P)+\bb)$ and the new public sphere network becomes  $\phi_U(\bZ;\btheta_U)=\phi_U'(\bW\bZ;\btheta_U)$, where $\phi_P'$ and $\phi_U'$ are the private and public sphere networks from $\textbf{CNN-MMD}$, respectively, and $\bW$ is an orthonormal matrix obtained by adding the penalty \eqref{eq:orthonormality} to the private sphere network objective. We found the orthonormality of $\bW$ to be crucial for this model to be able to perform the utility task. 
    \item \textbf{SCNN-KDI/WDN/LSDN}: We use the same private and public sphere networks as \textbf{NN-MMD}, but replace the privacy loss with one of the three alternatives described in Section \ref{subsubsec:Priv Obj}. For the privacy discriminators $\phi_D(\cdot;\btheta_D)$ of the WDN and LSDN models, we found that using dense networks leads to better results.\footnote{We believe that dense networks can adapt faster to the changes in the private sphere networks, which makes dense privacy discriminators easier to train than convolutional ones.} Thus, we use two hidden layers with $1024$ units each for the privacy discriminators.
    \item \textbf{SCNN-(RF)MMD}: The methodology is the same as \textbf{SCNN-MMD}, but the mixture of Gaussian kernels is replaced with a $1000$-dimensional Random Fourier approximation, as was performed in Section \ref{subsec:Learning Privacy Enhancing}. 
\end{itemize}

Comparisons of these models are displayed in Figure \ref{fig:yaleb_end1}. We see that \textbf{CNN-MMD} does not achieve desirable trade-offs, proving completely ineffective on YaleB data and having very limited success in hiding private information on MeGlass data. This is because \emph{a single CNN layer only performs local transformations of input features, making shallow CNNs incapable of performing utility maximizing and privacy preserving transformations that need to be global in scale}. Adding subspace projections between convolutional layers alleviates this problem, which is why \textbf{SCNN-MMD} achieves far more desirable utility/privacy trade-off curves. 

Comparing \textbf{SCNN-MMD} and \textbf{NN-MMD} reveals that adding convolutions before subspace projections can significantly improve the utility performances for all levels of privacy. On YaleB, a convolutional layer in the private sphere helps preserve more utility information as the privacy level increases, and on MeGlass, this addition improves the utility performance of the system regardless of the privacy level.

Comparing the four privacy objectives with each other, we see once again that MMD and KDI achieve similar performances with KDI having a slight edge. On YaleB, \textbf{SCNN-MMD} and \textbf{SCNN-KDI} are able to capture more utility information in high-privacy settings, which may be due to the privacy discriminators in \textbf{SCNN-WDN} and \textbf{SCNN-LSDN} having difficulty capturing all the information encoded by the private sphere network. On MeGlass, \textbf{SCNN-MMD}, \textbf{SCNN-KDI}, \textbf{SCNN-WDN} and \textbf{SCNN-LSDN} all achieve near ideal utility/privacy trade-off curves (small utility performance losses, while privacy performances are near the proportions of the majority classes), showing all the privacy objectives to be successful in this setting. Finally, we see that \textbf{SCNN-(RF)MMD} achieves similar privacy performances to \textbf{SCNN-MMD} while having a slightly lower performance on MeGlass. This displays that \emph{even though weaker kernels can be useful for defining privacy objectives, they can lead to more privacy leakage in some instances}.

\subsection{Summary}
We performed four experiments, two of which explore linear projections on mobile sensing data and two of which explore convolutional mappings on face data. These experiments reveal that it is possible and beneficial to optimize privacy enhancing representations of the data together with the predictors that use them as inputs. Moreover, we see that our objectives and training methods are viable for optimizing simple linear projections as well as more complicated neural network mappings.

Two of our experiments were performed on data, where the utility prediction task can be performed independently from the private information (user identity), namely, the HAR and MeGlass data. Our experiments on these showcase the ability of our systems to remove almost all the sensitive information while maintaining high utility performances. While the conditions on MHEALTH and YaleB are less favorable (by our design in YaleB's case), our methods, nonetheless, allow users to select a desired level of utility and privacy for the feature mappings they choose. 

Since the use of discriminator networks did not improve the results on any of the datasets, it seems practical to use the closed-form MMD and KDI statistics as the privacy objective functions in general. For discrete private variables (as in our experiments), we found MMD to be the most practical privacy objective function to utilize, and it compares favorably to the other objectives. KDI compares favorably to MMD even with discrete private variables. Therefore, KDI could be a good objective in applications with continuous private variables, though it has a higher computational cost associated with it (cubic in batch size as opposed to quadratic). We found on HAR and YaleB datasets that, although WDN and LSDN perform similarly to MMD and KDI in low privacy settings, they may lead to lower utility performances in high privacy settings. It might be possible to improve these objectives by exploring other discriminator and private sphere network architectures, but it is unclear whether this can lead to more desirable utility/privacy trade-offs compared to MMD and KDI with a generic mixture of Gaussian kernels.

\section{Discussion}

Controlling the usage of one's information is often not possible once that information is shared with outside parties, hence, preemptively desensitizing their data might be one of the best defenses users have against intrusive inferences. The methods in this paper are thus geared towards removing sensitive information from the data before the users abdicate control of them, even for a desired utility. To motivate this approach, we demonstrated the viability of our methods with a comprehensive set of optimization objectives and focused on computationally cheap feature maps.

While the intent of our work is to immediately benefit some applications on sensitive data, our treatment is far from covering an exhaustive set of data processing techniques. For instance, noise addition mechanisms could also be included in the processing before the data is released. Such a mechanism can be optimized through the parameters of another neural network, which transforms an independent noise distribution before it is added to the data within the private sphere. 

Although we only applied our methods to simple dense and convolutional neural network models, the presented measures are suitable for optimizing other network architectures such as recurrent neural networks. 
%For example, the privacy objectives could be applied after a series of recurrent neural network (RNN) transformations or between the layers of Densely Connected Convolutional Neural Networks \cite{huang2017densely} by utilizing the subspace projections in \eqref{eq:cnn_funnel}. 
Additionally, the subspace projection layers included in the CNN architectures provide a simple method for reconstructing images from their dense mappings, but a similar functionality could be achieved by auto-encoders placed at the intersection of the public and private spheres. Auto-encoders could also help desensitize the data in the absence of well-defined utility targets, with privacy objectives applied to their bottleneck layers. 

For designing systems that take into account various facets of user privacy, our compression methods would inevitably have to be combined with other privacy paradigms and techniques. It may be desirable to extend our current learning methodology to a federated learning approach, and/or use  Differential Privacy mechanisms during the training of our models so that a user's participation is not revealed by the learned feature mappings. One may also consider performing homomorphic encryption so that users can avoid revealing sensitive information to any party while training the models meant to protect them.  

\section{Conclusion}

We proposed highly flexible feature mappings and training objectives, which help sensitive information to be removed from data intended only for a set of utility tasks. Our methods optimize the privacy enhancing feature maps and predictive models simultaneously in an end-to-end fashion, which enables users to limit the information they share without sacrificing the benefits from useful data analysis. The privacy objectives we presented in this paper are comprehensive, and we hope that they will provide a solid baseline for future works into privacy enhancing machine learning.

\section*{Acknowledgment}
This work was supported in part by the Brandeis Program of the Defense Advanced Research Project Agency (DARPA) and Space and Naval Warfare System Center Pacific (SSC Pacific) under Contract No. 66001-15-C-4068.

\footnotesize
\bibliographystyle{IEEEtran}
\bibliography{refs}

\end{document}